\documentclass[10pt,twocolumn,letterpaper]{article}

\usepackage[algorithms]{wacv}      %

\usepackage{graphicx}
\usepackage{booktabs}
\usepackage{multirow}
\usepackage{amsmath}
\usepackage{amssymb}
\usepackage{stmaryrd}
\usepackage{subcaption}
\usepackage{stmaryrd}
\usepackage{wrapfig}
\usepackage{enumitem}
\usepackage{cuted}
\usepackage{bbm}
\usepackage[accsupp]{axessibility}  %

\usepackage[pagebackref,breaklinks,colorlinks]{hyperref}

\DeclareMathOperator*{\argmin}{arg\,min}

\usepackage[capitalize]{cleveref}
\crefname{section}{Sec.}{Secs.}
\Crefname{section}{Section}{Sections}
\Crefname{table}{Table}{Tables}
\crefname{table}{Tab.}{Tabs.}

\def\wacvPaperID{177} %
\def\confName{WACV}
\def\confYear{2025}

\title{\texttt{PICASSO}: A Feed-Forward Framework for Parametric Inference of CAD
Sketches via Rendering Self-Supervision}

\author{Ahmet Serdar Karadeniz$^{1}$
\qquad
Dimitrios Mallis$^{1}$
\qquad
Nesryne Mejri$^{1}$
\qquad
Kseniya Cherenkova$^{1,2}$\\
Anis Kacem$^{1}$
\qquad
Djamila Aouada$^{1}$\\
$^{1}$ SnT, University of Luxembourg \quad $^{2}$ Artec 3D
\qquad
}

\begin{document}

\maketitle

\begin{abstract}
This work introduces \texttt{PICASSO}, a framework for the parameterization of 2D CAD sketches from hand-drawn and precise sketch images. \texttt{PICASSO} converts a given CAD sketch image into parametric primitives that can be seamlessly integrated into CAD software. Our framework leverages \textit{rendering self-supervision} to enable the pre-training of a CAD sketch parameterization network using sketch renderings only, thereby eliminating the need for corresponding CAD parameterization. Thus, we significantly reduce reliance on parameter-level annotations, which are often unavailable, particularly for hand-drawn sketches. 
The two primary components of \texttt{PICASSO} are \textbf{(1)}~a \textit{Sketch Parameterization Network} (SPN) that predicts a series of parametric primitives from CAD sketch images, and \textbf{(2)}~a \textit{Sketch Rendering Network} (SRN) that renders parametric CAD sketches in a differentiable manner and facilitates the computation of a rendering (image-level) loss for self-supervision. We demonstrate that the proposed \texttt{PICASSO} can achieve reasonable performance even when finetuned with only a small number of parametric CAD sketches. Extensive evaluation on the widely used SketchGraphs~\cite{seff2020sketchgraphs} and \textit{CAD as Language}~\cite{ganin2021computer} datasets validates the effectiveness of the proposed approach on \textit{zero-} and \textit{few-shot} learning scenarios.

\end{abstract}

\section{Introduction}
\label{sec:intro}

\begin{figure}
\setlength{\belowcaptionskip}{-0.5cm}
    \centering
\includegraphics[width=\linewidth]{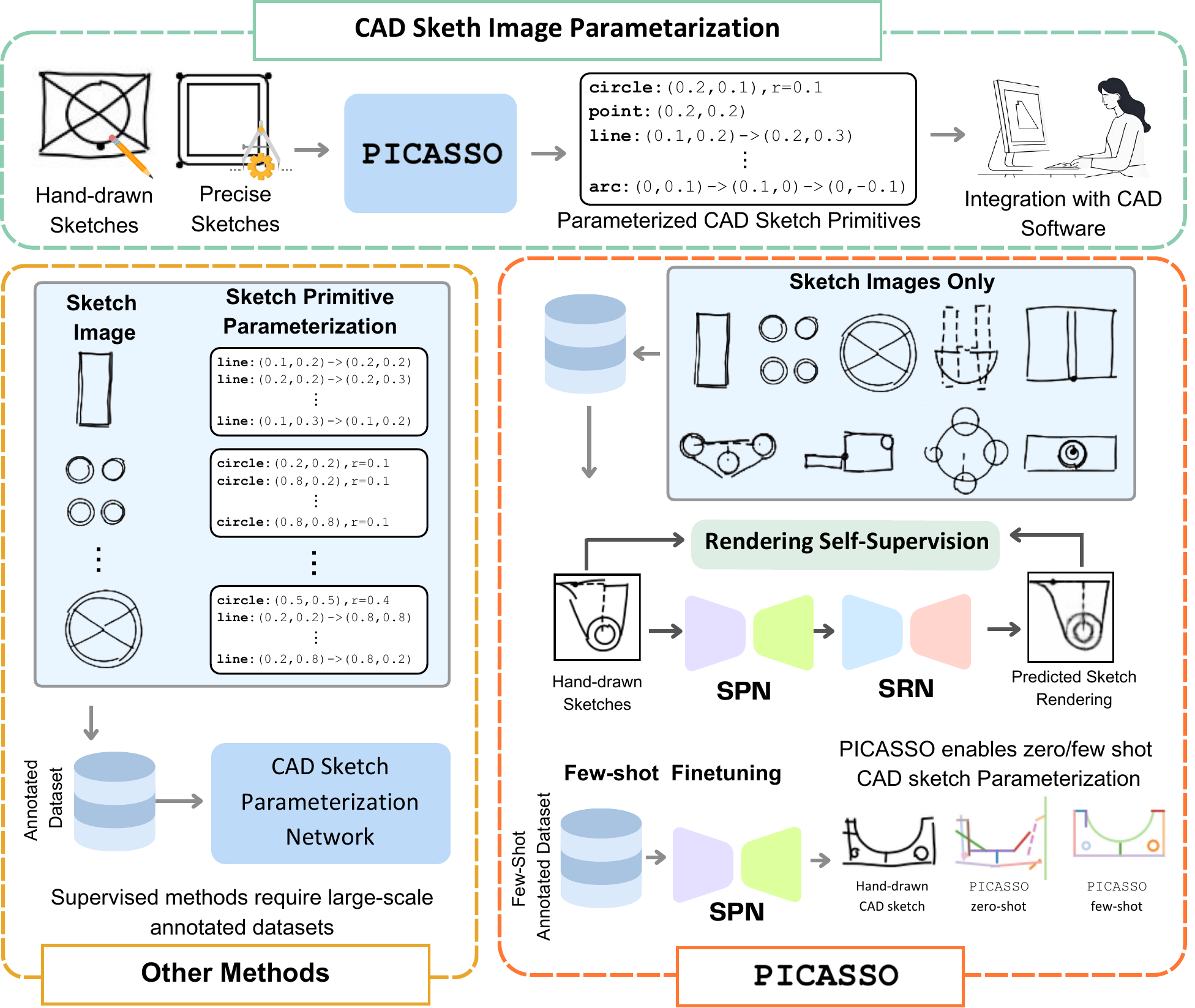}
\vspace{-0.5cm}
    \caption{\texttt{PICASSO} is a novel self-supervised framework for CAD sketch parameterization. Unlike methods relying on parametric annotations, \texttt{PICASSO} is pretrained via rendering self-supervision on CAD sketch images only, thus drastically reducing the need for parametrically annotated sketches. We demonstrate \texttt{PICASSO}'s effectiveness in both few-shot and zero-shot settings.}
    \label{fig:Teaser}
\end{figure}

Computer-Aided Design (CAD) has become the industry norm for mechanical design of any product prior to manufacturing. CAD software~\cite{Solidworks, Onshape} enhances the productivity of engineers and enables efficient extension or alteration of existing designs. The modern CAD workflow widely adopts the paradigm of \textit{feature-based} modeling~\cite{xu2021inferring}, where initially a series of two-dimensional parametric CAD sketches is specified, followed by CAD operations (\eg extrusion, revolution, etc.) to form a 3D solid. CAD sketches comprise a collection of geometric primitives (\eg lines, arcs) as well as constraints enforced between those primitives (\eg~coincident, parallel, etc.). Feature-based CAD constitutes an efficient way of constructing complex 3D models~\cite{khan2024cad, dupont2022cadops, mallis2023sharp, dupont2024transcad}. Commonly, the modelling process starts with
conceptualization of a design by roughly drawing it by hand~\cite{seff2022vitruvion}. 
Designers are tasked with meticulously translating these drawings, often in the form of raster images, into parametric CAD sketches.
Depending on the design complexity, the task can be time-consuming even for highly skilled designers. Thus, the automation of the CAD parameterization process (Figure~\ref{fig:Teaser}-\textit{top}) has gained attention in the research community~\cite{yang2022discovering,seff2022vitruvion} and CAD industry~\cite{cadCADSketchInternational,Solidworks}.

\begin{figure*}[t]
\setlength{\belowcaptionskip}{-0.5cm}
\begin{center}
\includegraphics[width=0.8\linewidth]{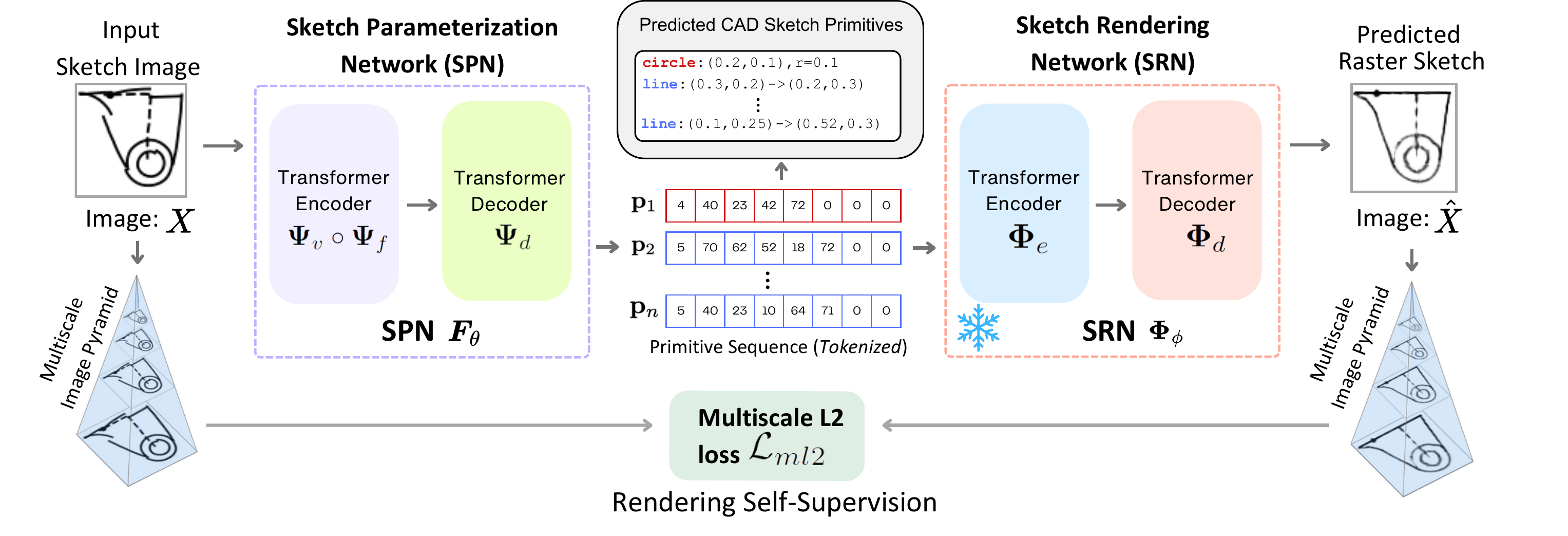}
\end{center}
\vspace{-0.5cm}
\caption{Framework overview. \texttt{PICASSO} is composed of two networks, namely the \textit{Sketch Parametrization Network} (SPN) and \textit{Sketch Rendering Network} (SRN). Once trained, SRN is kept frozen and used for rendering self-supervision using a multiscale $l2$ loss. This allows for image-level pre-training of the CAD sketch parameterization network SPN.  
}
\label{fig:method_intro}
\end{figure*}

Parameterization of CAD sketches from raster images constitutes a complex problem due to the large solution space required to model parametric entities, the nuanced intricacies of sketch designs, and the inherently inaccurate nature of hand-drawings. Compared to vectorized stroke inputs \cite{alaniz2022abstracting}, raster images pose additional challenges due to indistinguishable, overlapping or closely positioned primitives on the image domain. The availability of large-scale CAD sketch datasets~\cite{seff2020sketchgraphs, ganin2021computer} has enabled tackling this problem with learning-based approaches. Recent works~\cite{ganin2021computer,para2021sketchgen,seff2022vitruvion} propose \textit{autoregressive} generative models to learn CAD sketch parameterization through general-purpose language modelling. Similarly to next word prediction for natural language processing (NLP), these models predict the next primitive in the sketch and can be conditioned on images for sketch parameterization. Transformer-based networks are typically trained to predict tokens representing the type and parameters of each primitive via parameter-level supervision. These networks are adept at exploiting large-scale annotated datasets with millions of sketches (\eg SketchGraphs~\cite{seff2020sketchgraphs}) (Figure~\ref{fig:Teaser}-\textit{bottom left}). %
Nevertheless, there are real-world situations where this quantity of labeled sketches is unavailable. Examples include rough hand-drawn CAD sketches and 2D cross-sections~\cite{buonamici2018reverse} from 3D scans, where collecting parameter annotations is challenging. These examples exhibit a distribution shift \wrt to CAD sketches in large-scale datasets, underscoring the necessity of learning CAD sketch parametrization from raster images when parameter-level annotations are scarce or nonexistent.

In this work, we introduce a framework for \textbf{P}arametric \textbf{I}nference of \textbf{CA}D
\textbf{S}ketches via Rendering \textbf{S}elf-Supervisi\textbf{O}n, refereed to as \texttt{PICASSO}. \texttt{PICASSO} enables learning parametric CAD sketches directly from precise or hand-drawn images, even when parameter-level annotations are limited or unavailable (Figure~\ref{fig:Teaser}-\textit{bottom right}). This is achieved by utilizing the geometric appearance of sketches as a learning signal to pretrain a CAD parameterization network. Specifically, \texttt{PICASSO} is composed of two main components: \textbf{(1)} a \textit{Sketch Parameterization Network} (SPN) that predicts a set of parametric primitives from CAD sketch images and \textbf{(2)} a \textit{Sketch Rendering Network} (SRN) to render parametric CAD sketches in a differentiable manner. SRN enables image-to-image loss computation that can be used to pretrain SPN, leading to \textit{zero-} and \textit{few-shot} learning scenarios for hand-drawn sketch parametrization (see Figure~\ref{fig:Teaser}). To the best of our knowledge, we are the first to address CAD sketch parameterization with limited or without parametric annotations. \texttt{PICASSO} can achieve strong parameterization performance with only a small number of annotated samples. Moreover, departing from recent work based on autoregressive modelling~\cite{seff2022vitruvion,para2021sketchgen,ganin2021computer}, the proposed SPN is a \textit{feed-forward} network. The sketch is treated as an unordered set of primitives and a CAD sequence is predicted in a non-autoregressive manner. Experiments show that SPN outperforms recent autoregressive state-of-the-art on parameterizing precise and hand-drawn CAD sketch images. An overview of \texttt{PICASSO} is presented in Figure~\ref{fig:method_intro}.

\vspace{0.1cm}
\noindent \textbf{Contributions:} The main contributions of this work can be summarized as follows: 

\vspace{-0.15cm}
\begin{enumerate}[itemsep=0.05cm]
    \item \texttt{PICASSO} is a novel framework that enables image-level pretraining for CAD sketch parameterization, thus allowing, for the first time, few and zero-shot learning scenarios of CAD sketch parameterization from hand-drawn or precise CAD sketch raster images. 
    \item SRN, a neural differentiable renderer to rasterize CAD parametric primitives is proposed. By leveraging SRN, we are the first to explore image-level pretraining for CAD sketch parameterization.  
    \item The proposed feed-forward SPN results in state-of-the-art CAD sketch parameterization. 
    \item \texttt{PICASSO} is thoroughly evaluated both qualitatively and quantitatively on the widely-used \textit{SketchGraphs} dataset~\cite{seff2020sketchgraphs} under few or zero-shot evaluation settings. 
\end{enumerate}

\noindent This paper is organized as follows; Section~\ref{sec:related} reviews the related works. Section~\ref{sec:problem} formulates the problem of CAD sketch parameterization. The proposed \texttt{PICASSO} framework is described in Section~\ref{sec:method}. The experiments are provided in Section~\ref{sec:experiments}. Finally, Section~\ref{sec:conclusion} concludes this work.

\vspace{-0.5cm}
\section{Related Works}
\label{sec:related}

Most sketch-related literature focuses on understanding human-made sketches, typically represented as set of free-hand strokes. Relevant applications include sketch synthesis~\cite{sketchrnnICLR2018, AyanPixelor20}, recognition~\cite{li2015free,yu2017sketch}, segmentation~\cite{yang2021sketchgnn, qi2022one}, grouping of individual strokes~\cite{xu2022deep}, sketch classification~\cite{xu2022deep} and abstraction~\cite{alaniz2022abstracting}. Free-hand sketches are distinct from CAD sketches for feature-based CAD modelling. The free-hand representation is non-parametric, limited in terms of editability and generally is the result of a spontaneous drawing process. The focus of this work is the separate paradigm of CAD sketch parameterization~\cite{para2021sketchgen,seff2022vitruvion,seff2020sketchgraphs} from precise or hand-drawn CAD sketch images. The remainder of this section will introduce related methods for CAD parameterization or generation and discuss recent advancements in differentiable rendering of parametric entities.

\vspace{0.1cm} 
\noindent\textbf{CAD Sketch Parameterization:} CAD sketches relate to the formal profiling of mechanical components and are represented by a set of parametric entities (\eg, lines, circles, arcs, and splines) which are often constrained by defined relationships that maintain design intent. Typically the sketch is defined as a 2D drawing on a 3D plane, subsequently transformed into a 3D solid via CAD operations like extrusion, cutting, or revolution. Statistics from the Onshape CAD platform~\cite{Onshape} report that sketches constitute approximately 35\% of daily feature creation~\cite{onshapelink, ganin2021computer}. Literature on parametric CAD sketches mostly focuses on the task of CAD sketch generation or synthesis. Recent methods~\cite{ganin2021computer,para2021sketchgen,seff2022vitruvion} adopt a unified approach centered around autoregressive transformers. While the primary focus is on generation, transformer decoders can be further conditioned on sketch images for CAD parameterization. One of the first attempts to address CAD generation through generic language modelling was~\cite{ganin2021computer}. The authors employ the protocol buffer specification for 2D sketches and generate constrained CAD sketches using a transformer decoder. In \cite{willis2021engineering}, authors explore two distinct sketch representations based on either hypergraphs or turtle graphics (sequence of pen-up, pen-move actions) and introduce corresponding generative models, CurveGen and TurtleGen, respectively.

Concurrently, SketchGen~\cite{para2021sketchgen} and Vitruvion~\cite{seff2022vitruvion} propose two-stage architectures for both primitive and constraint generation. Seff~\etal~\cite{seff2022vitruvion} also investigate their model's efficacy for the parameterization of hand-drawn sketches. These methods follow the autoregressive learning strategy as it constitutes the natural choice for generative modelling. Autoregressive inference is also well-suited for the related application of sketch auto-completion but might be suboptimal for CAD parameterization from images (see detailed discussion in Section~\ref{sec:spn}). In contrast, the proposed SPN is a feed-forward network. Related to ours is also the non-autoregressive method of~\cite{yang2022discovering}. Authors introduce the task of modular concept discovery that is addressed through a program library induction perspective. Recently, the authors in~\cite{wang2024parametric} presented a concurrent work to ours. They highlighted the issue of error accumulation in autoregressive models and introduced a feed-forward strategy for CAD sketch parameterization, which is similar to our Sketch Parameterization Network (SPN). Also concurrently to ours, the work of ~\cite{karadeniz2024davinci} proposes a single-stage architecture for joint sketch parameterization and constraint prediction. All aforementioned methods solely rely on parameter-level supervision and overlook the rendered geometry of a CAD sketch. In this work, rendering self-supervision is proposed to provide an alternative signal for sketch parameterization. Such signal allows for pre-training CAD sketch parameterization network, hence enabling it to parameterize hand-drawn or precise CAD sketches under limited or without parametric supervision.

\vspace{0.1cm}
\noindent\textbf{Parametric Rendering:} Rendering refers to the process of converting vector parameters into a raster image. Given that direct vector supervision is not always available, rendering modules can bridge the vector and raster domains and enable gradient-based optimization. Applications include visually supervised parameterization~\cite{mo2021general, egiazarian2020deep,reddy2021im2vec, ma2022towards, chen2023editable, li2020differentiable}, as well as painterly rendering or seam carving~\cite{li2020differentiable}. Generally, parametric rendering cannot be directly integrated within end-to-end training pipelines, since vectorized shapes (represented as indicator functions) are not differentiable. A line of work explores differentiable renderers~\cite{huang2019learning,li2020differentiable} that can automatically compute gradients with respect to input vector parameters. Even though these methods are generalizable, gradients are only given for continuous parameters and cannot affect discrete decisions such as adding, rearranging or removing primitives~\cite{li2020differentiable}. More relevant to us, a line of work investigates learnable rendering approaches~\cite{huang2019learning, Nakano2019NeuralPA,Zheng2018StrokeNetAN,mo2021general} to convert parameters into raster images that allow optimization with image-base losses. Rendering modules proposed in these works operate on vector graphics, commonly in the form of Bezier paths. Compared to vector graphics, rendering used for self-supervision of parametric  CAD sketches should allow the capturing of multiple types of parametric primitives (\eg, lines, circles).  To our knowledge, we are the first to explore neural rendering of parametric CAD sketches.

\vspace{0.1cm}

\section{Problem Statement}
\label{sec:problem}

Given a binary sketch image $\mathbf{X} \in \{0,1\} ^{h \times w}$, where $h$ and $w$ denote the height and the width, respectively, our goal is to infer 
a set of $n$ parametric primitives 
\hbox{$\{\mathbf{p}_1, \mathbf{p}_2, ..., \mathbf{p}_n\} \in \mathcal{P}^n$}
reconstructing the input image $\mathbf{X}$. Here, $\mathcal{P}$ denotes the space of possible primitives. Similarly to \cite{seff2022vitruvion}, each $\mathbf{p}_i \in \mathcal{P}$ can be one of the following types:

\vspace{0.1cm}
\noindent
\textbf{Line}. A line $\mathbf{l}_i$ is defined by its start and end points \hbox{$(x_s, y_s) \in \mathbb{R}^2, (x_e, y_e) \in \mathbb{R}^2$}.

\noindent
\textbf{Circle}. A circle $\mathbf{c}_i$ is represented by its center point \hbox{$(x_c, y_c) \in \mathbb{R}^2$}, and its radius $r \in \mathbb{R}$.

\noindent
\textbf{Arc}. An arc $\mathbf{a}_i$ is defined by its start, middle, and end points $(x_s, y_s) \in~\mathbb{R}^2$, $(x_m, y_m)~\in~\mathbb{R}^2$, $(x_e, y_e) \in \mathbb{R}^2$.

\noindent
\textbf{Point}. A point $\mathbf{d}_i$, parameterized through its coordinates \hbox{$(x_p, y_p) \in \mathbb{R}^2$}. 
\vspace{0.1cm}

\begin{figure}[t]
    \setlength{\belowcaptionskip}{-0.5cm}
    \centering
    \includegraphics[width=\linewidth]{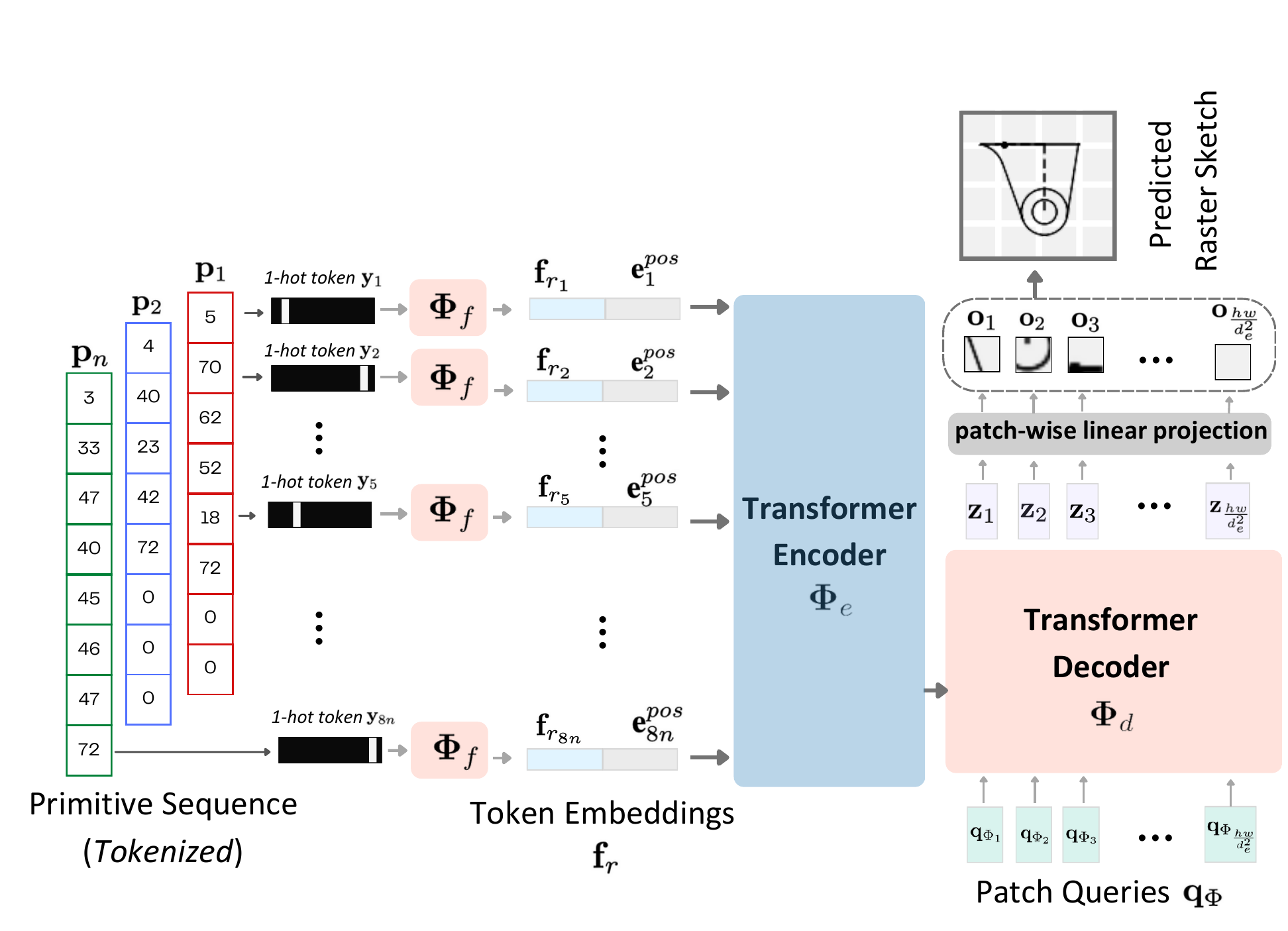}
    \caption{Overview of the \textit{Sketch Rendering Network} (SRN). SRN is modeled by a transformer encoder-decoder that learns a mapping from parametric primitive tokens to the sketch image domain. SRN enables neural differentiable rendering that we leverage for rendering self-supervision of SPN.}
    \label{fig:srn}
\end{figure}

Our goal becomes to learn the mapping \hbox{$\boldsymbol{F}_{\theta}: \{0,1\} ^{h \times w} \rightarrow \mathcal{P}^n$} with model parameters $\theta$, from the image domain to the space of parametric primitives. As in~\cite{seff2022vitruvion}, each primitive is represented by a set of \textit{tokens} defining primitive types and quantized primitive parameters. 
A $6$-bit uniform quantization is adopted. More specifically, every primitive is expressed as a set of 8~tokens $\mathbf{p}_n~=~\{t_i^j\}_{j \in \llbracket 1 .. 8 \rrbracket}$ with $t_i^j \in \llbracket 0..72 \rrbracket$.  Tokens in $\llbracket 0..6 \rrbracket$ represent primitive types \ie, \textit{padding}, \textit{start}, \textit{end}, \textit{arc}, \textit{circle}, \textit{line}, and \textit{point}, respectively. Tokens in the range $\llbracket 7..70 \rrbracket$ correspond to quantized primitive parameters and values $\{71,72\}$ refer to whether the primitive is a construction primitive (used by designers for referencing). The complete set of quantized primitives can be formally defined as ${\mathbf{y}}~\in~\{0,1\}^{8n \times 73}$ where $8$ represents the tokens for each primitive and $73$ is the tokenization interval.

\vspace{-0.1cm}
\section{\texttt{PICASSO}: Proposed Framework}
\label{sec:method}

\texttt{PICASSO} comprises two transformer-based networks, namely the Sketch Rendering Network (SRN) and the Sketch Parameterization Network (SPN). We leverage the differentiable neural renderer SRN for rendering self-supervision, enabling large-scale pre-training of CAD parameterization models without requiring any parameter-level annotations. Details on both SRN and SPN along with the CAD sketch parameterization scenarios enabled by the proposed image-level pre-training are described next.

\subsection{Sketch Rendering Network}
\label{sec:srn}

This work proposes a Sketch Rendering Network (SRN) designed for neural rendering of parametric CAD sketches. SRN learns $\mathbf{\Phi}_\phi:~\mathcal{P}^n~\rightarrow~\{0,1\}^{h \times w}$, \ie, the mapping from a set of parametric primitive tokens to the sketch image domain. The primary utility of SRN is to allow the computation of an image-to-image loss between predicted primitives and their corresponding precise or hand-drawn input sketch image, thus enabling image-level pre-training for CAD sketch parametrization. Note that \textit{explicit} rendering of CAD sketches implies the direct rasterization of parametric primitives, and it is inherently a non-differentiable process. While differentiable rendering solutions exist in literature~\cite{li2020differentiable}, their use is limited to parameter refinement. 

\vspace{0.1cm}
\noindent
\textbf{Network Architecture:} The proposed architecture for SRN is depicted in Figure~\ref{fig:srn}. SRN operates on one-hot input tokens ${\mathbf{y} = [\mathbf{y}_1 \;\mathbf{y}_2 \;...\; \mathbf{y}_{8n}]}~\in~\{0,1\}^{8n \times 73}$. Firstly, input tokens are individually projected through a linear layer $\mathbf{\Phi}_f$ to produce token embeddings  $\mathbf f_r = [ \mathbf{f}_{r_1},\mathbf{f}_{r_2}, ..., \mathbf{f}_{r_{8n}}]$, where $\mathbf f_{r_i} = \mathbf{\Phi}_f(\mathbf{y}_i)~\in~\mathbb R^{d_q}$ for the $i$-th token and $d_q$ denotes the embedding dimension. The $\mathbf f_r $ embeddings are concatenated with fixed positional embeddings $\textbf{e}^{pos}_{i}$~\cite{Parmar2018ImageT} and processed by a transformer encoder network $\mathbf{\Phi}_e$ without changing their dimensions. In the decoding phase, the decoder $\mathbf{\Phi}_d$, inspired by~\cite{he2022masked}, employs self and cross-attention to map a set of randomly initialized \textit{learnable} patch queries $\mathbf q_{\Phi}~\in~\mathbb{R} ^{\frac{hw}{d_e^2} \times d_e \times d_e}$ to a set of decoded patch embeddings $\mathbf{z} = [\mathbf{z}_1 \; \mathbf{z}_2 \;...\; \mathbf{z}_{\frac{hw}{d_e^2}}] ~\in~\mathbb{R} ^{\frac{hw}{d_e^2} \times d_e \times d_e}$. Patch queries $\mathbf q_{\Phi}$ serve as learnable tokens that query the encoded representations to extract spatial and structural information that is required for the rendering. Finally, each decoded patch embedding $\mathbf{z}_i$ pass through a patch-wise linear layer with a sigmoid activation function, to produce the corresponding output patch of pixel-intensities $\mathbf{o}_{\frac{hw}{d_e^2}}~\in [0,1]^{d_e \times d_e}$. The final rendered image is reconstructed by assembling the output patches $\mathbf{o}_i$.

\subsection{Sketch Parametrization Network}
\label{sec:spn}

\begin{figure*}[t]
    \setlength{\belowcaptionskip}{-0.5cm}
    \centering
    \vspace{-0.4cm}
    \includegraphics[width=.8\textwidth]{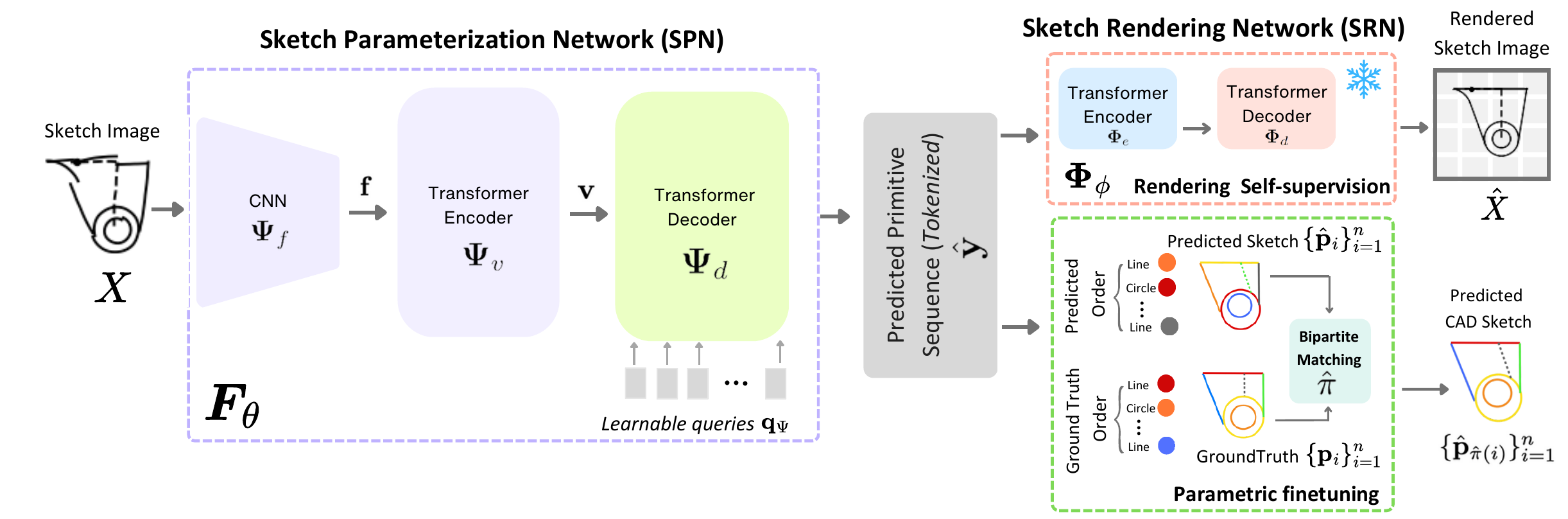}
    \caption{Overview of the \textit{Sketch Parameterization Network} (SPN). An input raster sketch image is processed by a convolutional backbone and the produced feature map is fed to a transformer encoder-decoder for sketch parameterization. SPN is pre-trained using rendering self-supervision provided by SRN, allowing zero-shot CAD sketch parameterization, and finetuned with parameter-level annotations for few-shot scenario.  
    }
    \label{fig:spn}
\end{figure*}

Our Sketch Parametrization Network (SPN) learns the inverse mapping $\boldsymbol{F}_{\theta}$ from the sketch image domain to the set of parametric token primitives (as defined in Section \ref{sec:problem}). 
$\boldsymbol{F}_\theta$ follows a feed-forward transformer encoder-decoder architecture inspired by~\cite{carion2020end}.

\vspace{0.1cm}
\noindent
\textbf{Network Architecture:} Figure~\ref{fig:spn} shows the architecture of SPN. First, a convolutional~\cite{Ronneberger2015UNetCN} backbone $\mathbf{\Psi}_f$ extracts a set of image features $\mathbf{f} \in \mathbb{R} ^{c \times h \times w}$, 
for input sketch image $\mathbf{X}$. The input feature map is divided into non-overlapping patches that are embedded along with fixed positional embeddings~\cite{Parmar2018ImageT} and passed through a standard vision transformer encoder~\cite{dosovitskiy2020image} $\mathbf{\Psi}_v$ to produce a set patches $\mathbf v~\in~\mathbb{R} ^{\frac{hw}{d_e^2} \times d_e \times d_e}$. The decoder $\mathbf{\Psi}_d$ is a vanilla transformer~\cite{vaswani2017attention} trained to decode a set of $d$ randomly initialized learnable query vectors $\mathbf q_{\Psi}~\in~~\mathbb{R}^{d\times d_q}$ 
into primitive tokens via self and cross-attention. Note that $d_e$ and $d_q$ maintain the same value as for SRN. The decoded tokens are classified independently via a linear layer with softmax activation resulting in token probabilities~$\hat{\mathbf{y}}~\in~[0,1]^{8n \times 73}$.    %

\vspace{0.1cm}
\noindent \textbf{Feed-Forward \vs Autoregressive Network Design:} The suggested learning function for CAD sketch parameterization $\boldsymbol{F}_\theta$, constitutes an unordered set learning strategy via feed-forward network design that departs from the sequence modelling achieved by autoregressive networks in recent works~\cite{ganin2021computer,para2021sketchgen,seff2022vitruvion}. Autoregressive methods are inherently order-aware as learning $\boldsymbol{F}_{\theta}$ results in next token prediction. 
Their main limitation is that the succession of primitives is clearly non-injective, and typically many primitive sequences might result in the same final geometry. Another strategy to mitigate order ambiguity is to sort primitives by parameter coordinates.
Both sorting and ordered sketch parameterization strategies greatly expand the possible solution space. Additionally, autoregressive inference may introduce inconsistencies such as exposure bias~\cite{schmidt2019generalization} that impair the model's learning ability. Thus, we propose a feed-forward strategy for CAD sketch parameterization.

\subsection{Rendering Self-Supervision for CAD Sketch Parameterization}
\label{sec:pretraining}

While parametric supervision can train well-performing CAD sketch parameterization models from raster images~\cite{seff2022vitruvion,ganin2021computer}, required annotations are not always available. Particularly for the parameterization of hand-drawn sketches, recent methods~\cite{seff2022vitruvion} are trained on synthesized hand-drawn samples to ensure the availability of parameter-level annotations. Nonetheless, this approach entails acquiring knowledge of the distribution specific to artificial hand-drawn sketches, potentially hindering the model's ability to generalize effectively to sketches created by humans.

\texttt{PICASSO} pre-trains a sketch parametrization network SPN via a neural renderer SRN. An input hand-drawn or precise sketch image $\mathbf X$ is fed to SPN, which in turn encodes it into a set of primitive tokens in $\hat{\mathbf{y}}~\in~[0,1]^{8n \times 73}$ that are rendered by SRN. The resulting raster sketch image $\mathbf{\hat{X}}$ along with the input image $\mathbf{{X}}$ are used within a \textit{multiscale} $l2$ loss similar to \cite{Reddy2021Im2VecSV}. Specifically, the rendered sketch image by SRN and the input raster image are successively downsampled to obtain multiscale image pyramids. The image loss is computed at each pyramid level as follows, 

\vspace{-0.5cm}
\begin{equation}
    \mathcal{L}_{ml2} =  \sum_{s \in S}  \| d_s\left(\mathbf{\Phi}_\phi(\boldsymbol{F}_\theta(\mathbf{\hat{X}})\right)) - d_s\left(\mathbf{X}\right) \|_2^2 \ ,
    \label{eq:img-loss}
\end{equation}
\vspace{-0.5cm}

\noindent where $d_s(.)$ represents the downsampling operation for a scale $s$ and the pyramid level is denoted by $S \in \llbracket 1..5 \rrbracket$. This mechanism ensures that if the rendered sketch only partially overlaps with the input raster image at a higher resolution, coarser resolutions can produce informative gradients for SPN.  Note that the SRN is first learned synthetically and subsequently kept frozen during the training of the CAD parameterization model, preventing it from shifting towards a non-interpretable latent space. 

\vspace{0.1cm}
\noindent \textbf{Synthetic Training of SRN:} Training of SRN requires a set of parametric primitives and their corresponding sketch images. In practice, we train it synthetically by randomly generating primitives and their explicit renderings. Given a a collection of sketch images $\mathcal{X} = \{\mathbf{X}_z\}_{z=1}^{N_z}$ along with token one-hot encoding $\{\mathbf{y}_z\}_{z=1}^{N_z}~\in~[0,1]^{8n \times 73}$, SRN is similarly learned through the multiscale $l2$ loss, formally $\mathcal{L}_{ml2} =  \sum_{s \in S}  \| d_s(\mathbf{\Phi}_\phi(\mathbf{y}_z)) - d_s(\mathbf{X}_z) \|_2^2$.

\subsection{Zero- and Few-Shot Learning of CAD Sketch Parameterization}
\label{sec:zero_few}

The pre-training described in Section~\ref{sec:pretraining} allows for CAD sketch parameterization from precise and hand-drawn sketch images when parametric annotations are limited or unavailable. For the zero-shot setting, the SPN model, pre-trained with rendering self-supervision is directly used to infer a set of parametric primitives $\{\mathbf{p}_1, \mathbf{p}_2, ..., \mathbf{p}_n\}$ from the sketch image $\mathbf{X}$. In the few-shot scenario, the pre-trained SPN is finetuned with few annotated samples using parameter-level supervision. Note that due to the unordered set modeling strategy via the feed-forward nature of SPN, the order of predicted primitives does not necessarily match that of ground truth ones. Hence, the finetuning of SPN with parameter-level supervision is enabled through optimal bipartite matching. For predicted primitives $\{\hat{\mathbf{p}}_{i}\}_{i=1}^n$, we recover the permutation $\hat{\pi} \in \Pi_n$ such as

\vspace{-0.3cm}
\begin{equation}
    \hat{\pi} = \argmin_{\pi \in \Pi_n} \sum_{i=1}^n \mathcal{L}_{param}(\mathbf{p}_i, \hat{\mathbf{p}}_{\pi(i)}) \ ,
    \label{eq:bpm}
\end{equation}
\vspace{-0.3cm}

\noindent where $\Pi_n$ is the space of all bijections from the set $\llbracket 1..n \rrbracket$ to itself. The assignment $\hat{\pi}$ can be efficiently computed through Hungarian matching~\cite{kuhn1955hungarian}. $\mathcal{L}_{parm}$ is a cross-entropy loss between predicted and ground truth tokens.

\section{Experiments}
\label{sec:experiments}

\begin{figure*}[t]
\setlength{\belowcaptionskip}{-0.5cm}
    \centering
\includegraphics[width=.85\linewidth]{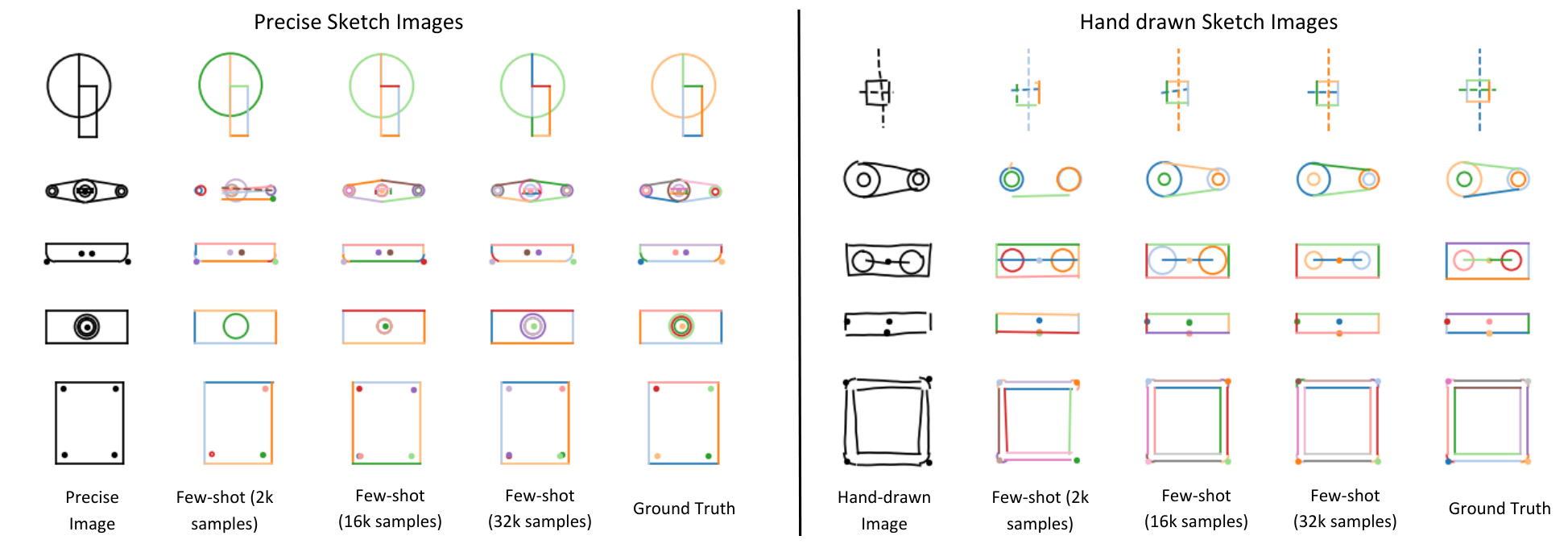}
    \vspace{-0.1cm}
    \caption{Few-shot setting. Qualitative results of \texttt{PICASSO} learned CAD sketch parameterization from precise and hand-drawn sketches.}
    \label{fig:qualitative-few}
\end{figure*}

\label{sec:details}

\noindent \textbf{Dataset:} We evaluate \texttt{PICASSO} on the SketchGraphs dataset~\cite{seff2020sketchgraphs} of parametric CAD sketches. We adopt the preprocessing of~\cite{seff2022vitruvion, para2021sketchgen} where duplicates and sketches containing less than $6$ primitives are removed. The final size of the dataset after filtering is around $1.53$ million sketches. We use the train/val split as in~\cite{seff2022vitruvion}. Our test partition includes $5000$ samples. We follow the hand-drawn synthesis strategy of~\cite{seff2022vitruvion} where primitives are subject to random translations/rotations and precise renderings are augmented with a Gaussian process model. A cross-dataset evaluation is also performed on the \textit{CAD as a Language} dataset~\cite{ganin2021computer}. We filter out samples that include splines and report performance on the first 5000 sketches of the test set.

\begin{figure}[h]
\setlength{\belowcaptionskip}{-0.5cm}
    \begin{subfigure}[t]{\linewidth}
        \begin{subfigure}{.45\linewidth}
            \centering
            \includegraphics[height=4cm,width=4cm,]{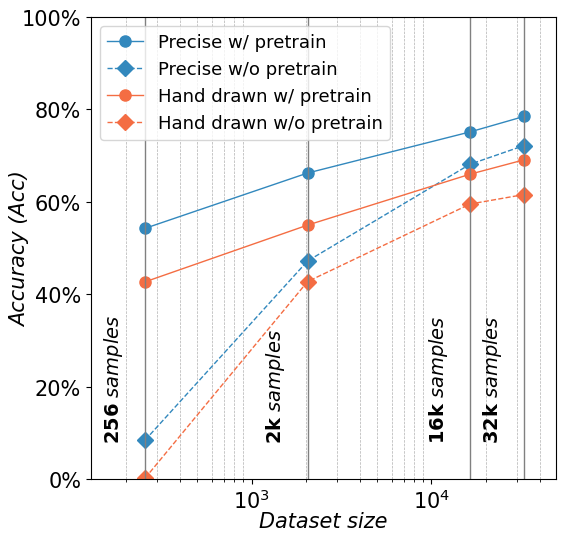}
        \end{subfigure}
        \begin{subfigure}{.45\linewidth}
            \centering
            \includegraphics[height=4cm,width=4cm]{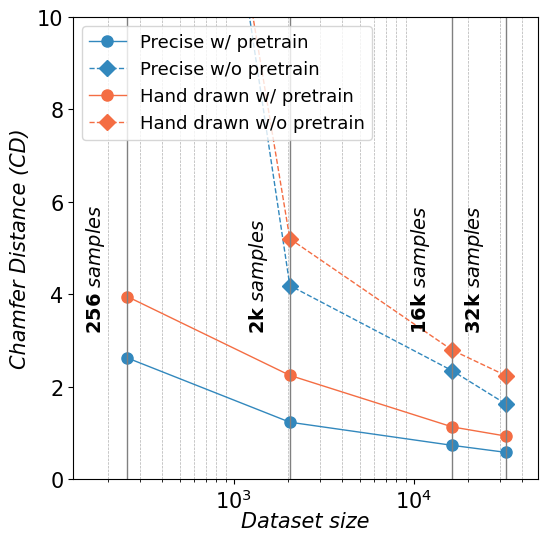}
        \end{subfigure}
    \end{subfigure}
    \vspace{-0.7cm}
    \caption{Effectiveness of rendering self-supervision on a few-shot evaluation. \texttt{PICASSO} is pre-trained on a dataset of both precise and hand-drawn sketches. We report (\textbf{left}) parametric \textit{Accuracy (Acc)} and (\textbf{right}) image-level \textit{Chamfer Distance (CD)} . Pre-trained SPN (w/ pre-train)  is compared to a from-scratch counterpart (w/o pre-train). Best viewed in colors.}\label{ref:facialresults}
    \label{fig:fewshot}
\end{figure}

\vspace{0.1cm}
\noindent \textbf{Implementation Details:} Images are of size $128 \times 128$. SPN uses a U-Net~\cite{Ronneberger2015UNetCN} convolutional backbone with a ResNet34~\cite{He2015DeepRL} that produces a $16 \times 128 \times 128$ feature map. For SPN, the transformer encoder-decoder is formed by $4$ layers each, with $8$ heads and $d_q=256$ latent dimensions. The SRN transformer has $12$ layers with $8$ heads and the same latent dimension as SPN. Patch size $d_e$ is set to $16$. As in~\cite{seff2022vitruvion}, the maximum number of primitives is fixed to $n=16$. We train SRN and SPN for $40$ and $20$ epochs respectively, with a learning rate of $1\times10^{-4}$ and a batch size of $128$. Note that SRN is trained with one-hot encoded primitive tokens as input but operates on token probabilities output by SPN for image-level pre-training. Primitives with incorrect syntax are removed. We implemented modules in Pytorch~\cite{Paszke2017AutomaticDI} and use the Adam optimizer~\cite{Kingma2014AdamAM}.

\vspace{0.1cm}
\noindent \textbf{Evaluation:} For quantitative evaluation, we report both \textit{parameter-based} and \textit{image-based} metrics. Parameter-based metrics are computed between predicted and ground truth primitive tokens. Note that \texttt{PICASSO} predicts primitives as an unordered set, so correspondence \wrt the ground truth is recovered through optimal bipartite matching as described in Eq.(\ref{eq:bpm}). To ensure a fair comparison, the same correspondence recovery step is performed for the autoregressive method of~\cite{seff2022vitruvion}. We report \textit{accuracy} and parametric \textit{mean-squared error (ParamMSE}). Accuracy considers the whole predicted sequence and ParamMSE is computed on the parameters of predicted tokens. Image-based metrics are reported on the explicit rendering of predicted primitive sequences. We measure a normalized pixel-wise \textit{Mean Squared Error} (\textit{MSE}) and bidirectional \textit{Chamfer Distance} (\textit{CD})~\cite{wu2021density}. To compute CD, a set of points is uniformly sampled on foreground pixel coordinates of the explicitly rendered sketches.

\subsection{Few-shot and Zero-shot CAD Sketch Parameterization}

\begin{table}
    \setlength{\belowcaptionskip}{-0.3cm}
    \setlength{\tabcolsep}{2pt}

    \resizebox{1.02\linewidth}{!}{
    \begin{tabular}{lcccccccc}
        \toprule
        \multicolumn{1}{c}{} & \multicolumn{4}{c}{\textbf{Precise Sketch Images}} & \multicolumn{4}{c}{\textbf{Hand-drawn Sketch Images}} \\
        \cmidrule(r){2-5} \cmidrule(l){6-9}
        Method & \textit{Acc} & \textit{ParamMSE} & \textit{ImgMSE} & \textit{CD} & \textit{Acc} & \textit{ParamMSE} & \textit{ImgMSE} & \textit{CD} \\
        \midrule
        Resnet34 & 0.465 & 908 & 0.199 & 5.883 &  0.396& 1048 & 0.240 & 6.908\\
        PpaCAD~\cite{wang2024parametric} & 0.524  & 589 & 0.195 & 5.097 & 0.464 & 744 & 0.244 & 6.904\\
        Vitruvion~\cite{seff2022vitruvion} & 0.537 & 624 & 0.186 & 4.901 &  0.461 & 685 & 0.237 & 5.258 \\
        \texttt{PICASSO} (w/o pt.) & 0.681  & 326 & 0.134 & 2.344 &   0.595 & 451 & 0.156 & 2.789  \\
        \textbf{\texttt{PICASSO} (w/ pt.) }& \textbf{0.751}  & \textbf{281} & \textbf{0.075}  & \textbf{0.729} &   \textbf{0.658} & \textbf{365} & \textbf{0.117} & \textbf{1.090}  \\
        \bottomrule

    \end{tabular}
    }
    \vspace{-0.2cm}
    \caption{Comparison with Vitruvion~\cite{seff2022vitruvion} and Resnet34, trained on $16k$ samples from SketchGraphs dataset~\cite{seff2020sketchgraphs}. Performance for \texttt{PICASSO} is reported w/o and w/ self-supervised pre-training. }
    \label{tab:fewshot}
\end{table}

\begin{table}
    \setlength{\belowcaptionskip}{-0.4cm}
    \setlength{\tabcolsep}{2pt}
    \resizebox{1.02\linewidth}{!}{
    \begin{tabular}{lcccccccc}
        \toprule
        \multicolumn{1}{c}{} & \multicolumn{4}{c}{\textbf{Precise Sketch Images}} & \multicolumn{4}{c}{\textbf{Hand-drawn Sketch Images}} \\
        \cmidrule(r){2-5} \cmidrule(l){6-9}
        Method & \textit{Acc} & \textit{ParamMSE} & \textit{ImgMSE} & \textit{CD} & \textit{Acc} & \textit{ParamMSE} & \textit{ImgMSE} & \textit{CD} \\
        \midrule
        Resnet34 & 0.524 & 829 & 0.189 & 5.698 &  0.448 & 946 & 0.230 & 6.692\\
        PpaCAD~\cite{wang2024parametric} & 0.562 & 601 & 0.272 & 6.600 & 0.508 & 754 & 0.289 & 8.193\\
        Vitruvion & 0.560 & 608 & 0.190 & 5.568 &  0.483 & 664 & 0.239 & 5.818 \\
        \textbf{\texttt{PICASSO}} & \textbf{0.809}  & \textbf{199} & \textbf{0.067} & \textbf{0.739} &  \textbf{0.669}  & \textbf{360} & \textbf{0.120} & \textbf{1.715} \\
        \bottomrule
    \end{tabular}
    }
    \vspace{-0.2cm}
    \caption{Cross-dataset few-shot evaluation on \textit{CAD as a Language}~\cite{ganin2021computer} dataset. Methods are trained on $16k$ samples.}
    \label{tab:fewshot_cad}
\end{table}

The proposed \texttt{PICASSO} enables the pre-training of CAD sketch parameterization directly from raster sketch images via rendering self-supervision. In this section, we evaluate the effectiveness of self-supervised pre-training for few and zero-shot settings. For results on hand-drawn sketch images, we follow \cite{seff2022vitruvion} and experiment on synthetic sketches to ensure data availability.%

\vspace{0.1cm}
\noindent\textbf{Few-shot Evaluation:} For the few-shot setting, we first pre-train \texttt{PICASSO} with rendering self-supervision (defined in Eq.(\ref{eq:img-loss})). The learned CAD sketch parameterization model is subsequently fine-tuned on smaller, curated sets of parameterized sketches (as discussed in Section.~\ref{sec:zero_few}). Note that two separate models are pre-trained and finetuned for hand-drawn and precise sketch images. In Figure~\ref{fig:fewshot}, the pre-trained \texttt{PICASSO} (w/ pt.) is compared to its from-scratch counterpart (w/o pt.). Overall, the pre-training outperforms learning from scratch across different sizes of fine-tuning datasets both with precise and hand-drawn images. A comparison with Vitruvion~\cite{seff2022vitruvion}, PpaCAD~\cite{wang2024parametric}, and ResNet34 baseline on a 16k-shot setting is reported in Table~\ref{tab:fewshot}. We observe that both Vitruvion and PpaCAD completely underperform when trained with only a small number of annotated samples. Furthermore, we demonstrate in \textcolor{black}{Figure~\ref{fig:qualitative-few}~(\textbf{right}) that \texttt{PICASSO} is able to to parameterize challenging hand-drawn sketches even when fine-tuned with a set of only $2$k annotated samples}. To assess the generalization capabilities of all methods, a cross-dataset evaluation is considered, where few-shot models trained on Sketchgraphs are tested on the CAD as a Language dataset~\cite{ganin2021computer}. Table~\ref{tab:fewshot_cad} shows that \texttt{PICASSO} consistently outperforms Vitruvion~\cite{seff2022vitruvion}, PpaCAD~\cite{wang2024parametric}, and the Resnet34 baseline.

\begin{figure}
\setlength{\tabcolsep}{1pt}
\setlength{\belowcaptionskip}{-0.3cm}
\centering
\begin{tabular}{@{}p{0.3\linewidth}@{} p{0.72\linewidth}@{}}
    \vspace{6pt} %
    \resizebox{0.95\linewidth}{!}{
    \begin{tabular}{lcc}
    \toprule
    Method & \textit{ImgMSE} & \textit{CD} \\ \midrule
    PMN~\cite{alaniz2022abstracting} & 0.233 & 3.243 \\
    \textbf{\texttt{PICASSO} } & \textbf{0.184} & \textbf{1.880} \\ 
    \bottomrule
    \end{tabular}
    } & 
    \vspace{0pt} %
    \includegraphics[width=\linewidth]{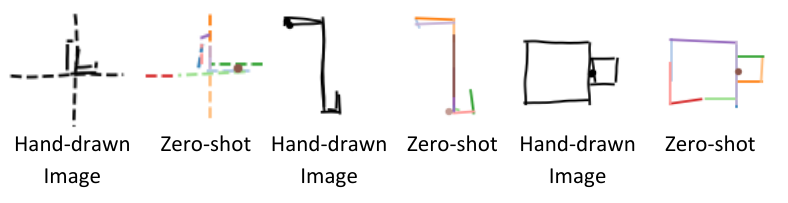} \\
\end{tabular}
\vspace{-0.5cm}
    \caption{Zero-shot evaluation on hand-drawn images. (\textbf{left}) Comparison \wrt the method of \cite{alaniz2022abstracting} adapted for CAD sketch parameterization and trained on the SketchGraphs dataset. (\textbf{right}) Qualitative results for \texttt{PICASSO} learned in a zero-shot setting.}
\label{tab:zeroshot}
\end{figure}

\vspace{0.1cm}
\noindent\textbf{Zero-shot Evaluation:} By leveraging rendering self-supervision, \texttt{PICASSO} can estimate the parameters of sketches directly without requiring parametric supervision. We evaluate the performance of \texttt{PICASSO} on the challenging zero-shot CAD sketch parameterization scenario in Figure~\ref{tab:zeroshot} (\textbf{left}). Comparison is performed \wrt to PMN~\cite{alaniz2022abstracting}, a recent self-supervised method for vectorized sketch abstraction. PMN operates on strokes (not images) that we form by sampling points on primitives. To adapt PMN for CAD sketch parameterization, we parameterize the predicted drawing primitive (stroke and type) via least-squares fitting. The network is trained on the SketchGraphs dataset~\cite{seff2020sketchgraphs}. We observe that \texttt{PICASSO} surpasses PMN performance in terms of image-based metrics. Note that we do not report \textit{Acc} and \textit{ParamMSE} on the zero-shot setting as any given CAD sketch can be constructed by arbitrarily many parameterizations and a self-supervised method cannot exactly match the parameters employed by the designer (by merging co-linear lines, flipping start-end points, etc.). Figure~\ref{tab:zeroshot} (\textbf{right}) shows some qualitative results of \texttt{PICASSO} on the zero-shot setting. Note that despite the large complexity of the task, \texttt{PICASSO} still recovers plausible sketch parameterization.

\subsection{Ablation Study}

\begin{table}[t]
\setlength{\belowcaptionskip}{-0.5cm}
\centering
\setlength{\tabcolsep}{3pt}
 \resizebox{0.4\linewidth}{!}{
\begin{tabular}{lcc}
\toprule
Method & \textit{ImgMSE} & \textit{CD}
\\ \toprule
Initial   & 0.047 & 0.24 \\
SPN + Diffvg &  0.087 & 0.48 \\
\textbf{\texttt{PICASSO}} & \textbf{0.045}  & \textbf{0.21}
\\ \bottomrule
\end{tabular}
}
\resizebox{0.59\linewidth}{!}{
    \begin{tabular}{lc|cc}
        \toprule
        \multicolumn{1}{l}{\textbf{Precise}} & \multicolumn{1}{c}{\textbf{Zero Shot}} & \multicolumn{2}{c}{\textbf{Few Shot (16k)}} \\
        \cmidrule(lr){2-2} \cmidrule(lr){3-4}
        Method & \textit{CD} $\downarrow$ & \textit{Acc} $\uparrow$ & \textit{CD} $\downarrow$ \\
        \midrule
        SPN $+$ DiffVG & 5.84 & 0.47 & 2.49\\
        \textbf{\texttt{PICASSO}} &\textbf{ 1.28} & \textbf{0.73} & \textbf{1.64}\\
        \bottomrule
    \end{tabular}
}

\vspace{-0.1cm}
\caption{ SRN-\texttt{PICASSO} is compared to the differentiable renderer DiffVG~\cite{li2020differentiable}. \textbf{(left)} Test-time optimization setting. \textbf{(right)} End-to-end training comparison. }
\label{tab:results_diffvg}
\end{table}

\noindent\textbf{SRN vs DiffVG:} We start by comparing the proposed SRN to the differentiable rendered DiffVG~\cite{li2020differentiable} on a test-time optimization setting. In particular, rendering self-supervision by both SRN and DiffVG is used to enhance CAD parameterization produced by a parameterically supervised SPN at test-time. This is similar to the widely used refinement of Scalable Vector Graphics (SVG), where differentiable renderers, such as DiffVG~\cite{li2020differentiable}, are commonly used to optimize predicted SVG parameters for an input sketch image. Table~\ref{tab:results_diffvg} \textbf{(left)} reports the results after optimization. DiffVG fails to optimize the sketch parameters and can even make the initial prediction less favourable. On the contrary, the proposed SRN improves the initial prediction in terms of image-based metrics. In Figure~\ref{fig:results_diffvg}, qualitative results of the test-time optimization are illustrated. We also compare SRN to DiffVG in the pretraining setting. While SPN can also be pretrained with DiffVG, we observed the following issues: (1) DiffVG is sensitive to SPN network size and primitive thickness due to sparsity of gradients~\cite{li2020differentiable}. It remains stable only with a small SPN ($1$-layer transformer) with $4 \times$ thicker primitives, whereas the original setting leads to degenerate geometries. (2) Convergence is slow, further hindered by DiffVG's batching inability\footnote{Issues \#17 and \#67 on \href{https://github.com/BachiLi/diffvg}{https://github.com/BachiLi/diffvg}}. As shown in Table~\ref{tab:results_diffvg} \textbf{(right)}, a small SPN pretrained with SRN surpasses DiffVG-based pretaining in zero/few-shot settings.

\vspace{0.1cm}
\noindent\textbf{SRN Rendering Performance:} Table~\ref{tab:results_srn_ablation} ablates different losses used to train SRN for parametric rendering. The quality of predicted raster images is evaluated \wrt ground truth explicit renderings. The multiscale $l2$ loss achieves the best performance on reported image-based metrics. 

\begin{table}[t]
\centering
\setlength{\tabcolsep}{8pt}
\setlength{\belowcaptionskip}{-0.3cm}
 \resizebox{0.65\linewidth}{!}{
\begin{tabular}{lcc}
\toprule
Loss Type & \textit{ImgMSE} & \textit{CD}
\\ \toprule
Binary Cross Entropy & 0.021 & 0.116 \\
$l2$ & 0.022 & 0.119 \\
\textbf{Multiscale $l2$ }& \textbf{0.020} & \textbf{0.109} \\
\bottomrule
\end{tabular}
}
\vspace{-0.2cm}
    \caption{Effect of different loss functions on SRN training.}
\label{tab:results_srn_ablation}
\end{table}

\begin{figure}[t]
\setlength{\belowcaptionskip}{-0.5cm}
    \centering
\includegraphics[width=\linewidth]{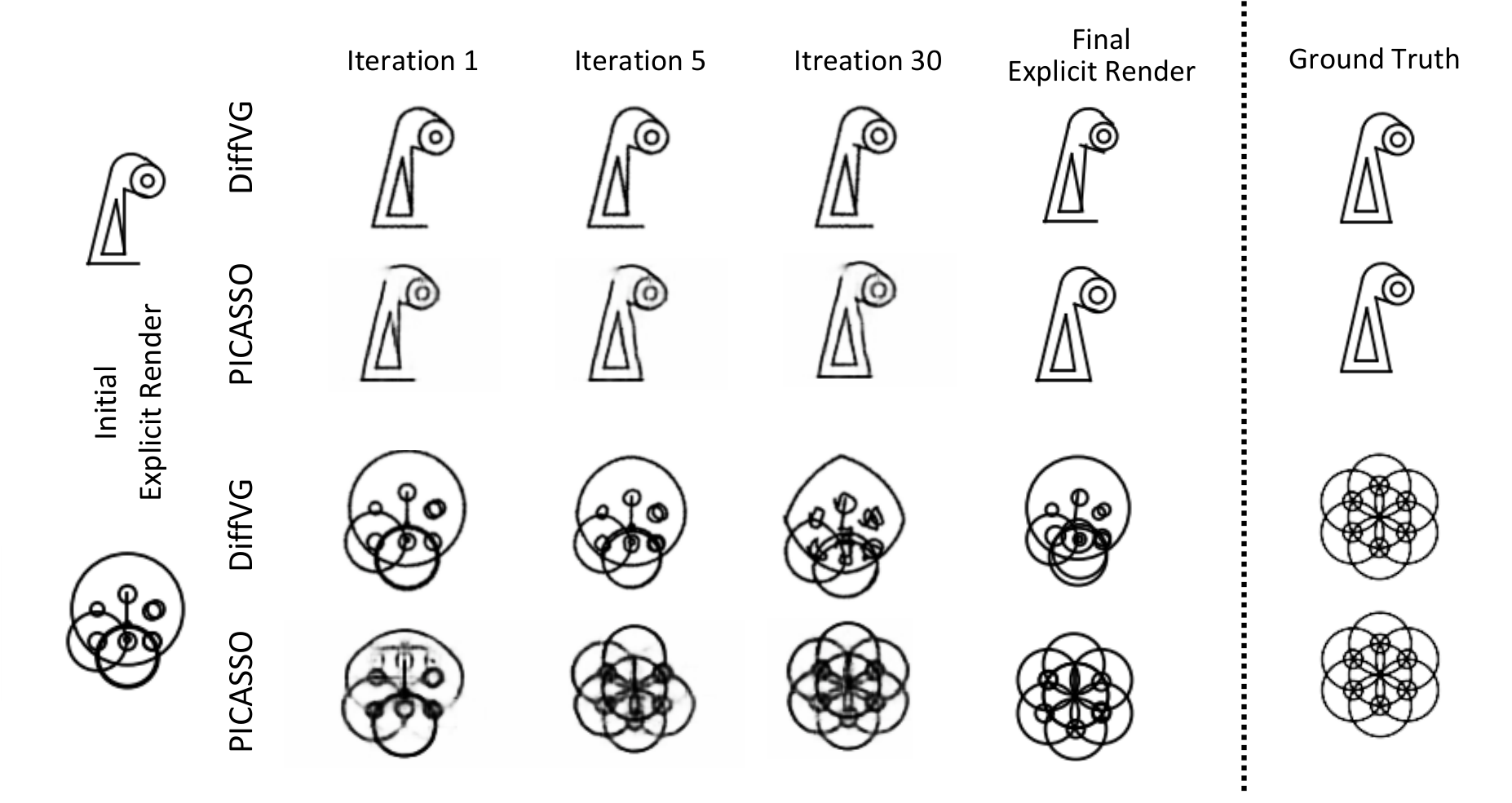}
    \vspace{-0.5cm}
    \caption{Test-time optimization with SRN-\texttt{PICASSO}. Comparison with DiffVG~\cite{li2020differentiable}. SRN-\texttt{PICASSO} progressively improves the parameterization and results in smooth transitions.}
    \label{fig:results_diffvg}
\end{figure}

\vspace{0.1cm}
\noindent\textbf{SPN with Parametric Supervision:} We evaluate SPN of \texttt{PICASSO} for CAD sketch parameterization of precise and hand-drawn sketch images. For this evaluation, SPN is trained with parametric supervision on the complete SketchGraphs dataset. Results are reported in Table~\ref{tab:results_vitruvion}. SPN surpasses both the autoregressive Vitruvion~\cite{seff2022vitruvion} and the non-autoregressive Resnet34 baseline by a large margin. It is also important to mention that SPN is approximately $10 \times$ faster than Vitruvion thanks to its feed-forward nature.

\begin{table}
\setlength{\belowcaptionskip}{-0.2cm}
    \centering
    \setlength{\tabcolsep}{3pt}
    \resizebox{1.02\linewidth}{!}{
    \begin{tabular}{lcccccccc}
        \toprule
        \multicolumn{1}{c}{} & \multicolumn{4}{c}{\textbf{Precise Sketch Images}} & \multicolumn{4}{c}{\textbf{Hand-drawn Sketch Images}} \\
        \cmidrule(r){2-5} \cmidrule(l){6-9}
        Method & \textit{Acc} & \textit{ParamMSE} & \textit{ImgMSE} & \textit{CD} & \textit{Acc} & \textit{ParamMSE} & \textit{ImgMSE} & \textit{CD} \\
        \midrule
        Resnet34 & 0.702 & 433 & 0.146 & 2.842 & 0.628 & 565 & 0.167 & 3.531\\
        Vitruvion & 0.813 & 226 & 0.052 & 0.267 & 0.658 & 391 & 0.112 & 0.875 \\
        \textbf{SPN} & \textbf{0.878} & \textbf{107} & \textbf{0.047} &  \textbf{0.237} & \textbf{0.827} & \textbf{146} & \textbf{0.075} & \textbf{0.499} \\
        \bottomrule

    \end{tabular}
    }
    \vspace{-0.2cm}
    \caption{Evaluation of the proposed feed-forward architecture on the SketchGraphs dataset~\cite{seff2020sketchgraphs}. For all methods, separate models are trained to parameterize precise and hand-drawn sketch images. }  %
\label{tab:results_vitruvion}
\end{table}

\begin{table}
\centering
\setlength{\belowcaptionskip}{-0.5cm}
\setlength{\tabcolsep}{3pt}
 \resizebox{0.56\linewidth}{!}{
            \begin{tabular}{lcccc}
            \toprule
                Method & \textit{Acc} & \textit{ParamMSE} & \textit{ImgMSE} & \textit{CD} \\
                \toprule
                {Sorted} & 0.722 & 432 & 0.114 & 2.20 \\
                {BM} & 0.831 & 165 & 0.053 & 0.33 \\
                \textbf{{BM}+UNet} & \textbf{0.878} & \textbf{107} & \textbf{0.047} & \textbf{0.23} \\
                \bottomrule
            \end{tabular}
}
 \resizebox{0.42\linewidth}{!}{
            \begin{tabular}{lcc}
                       \toprule
                Method & Precise & Hand-drawn\\
                \toprule
 \multicolumn{2}{l}{$\text{GTParams}_{\text{(upper bound)}}$} & 0.90 \\
 \midrule
                Vitruvion & 0.62 & 0.59  \\
                \textbf{SPN} & \textbf{0.64} & \textbf{0.62} \\
                \bottomrule
            \end{tabular}
}
\vspace{-0.2cm}
\caption{\textbf{(left)} Ablation on the various architectural components of SPN-\texttt{PICASSO}. We assess the impact of Bipartite Matching (BM) and the UNet-based backbone on SPN. \textbf{(right)} Performance for the constraint prediction of \cite{seff2022vitruvion} with input parameterization by SPN and~\cite{seff2022vitruvion} in terms of next-token accuracy. $\text{GTParams}$ represents the upper bound achieved with the ground truth primitives.}
\label{tab:spn_ablation}
\end{table}

\noindent\textbf{SPN Architecture:} Table~\ref{tab:spn_ablation} \textbf{(left)} depicts the ablation study of the architectural components of SPN. A transformer trained on sorted primitives is contrasted to a transformer learned with bipartite matching. Bipartite matching introduces a significant performance gain as the network does not allocate learning capacity on capturing primitive order. Finally, it is shown that the UNet-like backbone $\mathbf{\Psi}_f$ enhances the performance via multiscale feature extraction.

\vspace{0.1cm}
\noindent\textbf{Impact of CAD Sketch Parameterization on CAD Constraint Inference:} Parameterized primitives inferred by \texttt{PICASSO} can be directly constrained by existing constraint prediction models like \cite{seff2022vitruvion}. In Table~\ref{tab:spn_ablation} \textbf{(right)}, we report performance for the constraint prediction model of \cite{seff2022vitruvion} for input CAD parameterization predicted by SPN and \cite{seff2022vitruvion}. Our model improves constraint prediction performance by providing better CAD sketch parameterization.

\vspace{-0.2cm}
\section{Conclusions and Future Works}
\label{sec:conclusion}

In this paper, \texttt{PICASSO} is introduced, a novel framework for CAD sketch parameterization. It includes a feed-forward Sketch Parameterization Network (SPN) and a Sketch Rendering Network (SRN). SRN is a neural differentiable renderer that enables the use of rendering self-supervision for large-scale CAD sketch parameterization without requiring parametric annotations. This in turn accounts for real-world scenarios where large annotated hand-drawn CAD sketch datasets are not available. Experiments on few- and zero-shot settings validate the effectiveness of the proposed framework. Due to data unavailability, \texttt{PICASSO} was evaluated on four types of primitives (lines, points, arcs, and circles). In the supplementary, preliminary experiments are conducted with synthetic b-splines suggesting the applicability of \texttt{PICASSO} to free-form curves. This investigation in the context of CAD is left for the future.

\vspace{-0.3cm}
\section{Acknowledgements}
\noindent This work is supported by the National Research Fund (FNR), Luxembourg, BRIDGES2021/IS/16849599/FREE-3D project and by Artec3D.

{\small
\bibliographystyle{ieee_fullname}
\bibliography{main}

\begin{thebibliography}{10}\itemsep=-1pt

\bibitem{cadCADSketchInternational}
{C}{A}{D}{S}ketch - {C}{A}{D} {I}nternational --- cad.com.au.
\newblock \url{https://cad.com.au/software/cadsketch/}.
\newblock [Accessed 15-07-2024].

\bibitem{Onshape}
Onshape.
\newblock \url{https://www.onshape.com}.

\bibitem{Solidworks}
Solidworks.
\newblock \url{https://www.solidworks.com}.

\bibitem{alaniz2022abstracting}
Stephan Alaniz, Massimiliano Mancini, Anjan Dutta, Diego Marcos, and Zeynep Akata.
\newblock Abstracting sketches through simple primitives.
\newblock In {\em ECCV}, 2022.

\bibitem{AyanPixelor20}
Ayan~Kumar Bhunia, Ayan Das, Umar~Riaz Muhammad, Yongxin Yang, Timothy~M. Hospedales, Tao Xiang, Yulia Gryaditskaya, and Yi-Zhe Song.
\newblock Pixelor: A competitive sketching ai agent. so you think you can sketch?
\newblock {\em ACM TOG}, 2020.

\bibitem{buonamici2018reverse}
Francesco Buonamici, Monica Carfagni, Rocco Furferi, Lapo Governi, Alessandro Lapini, and Yary Volpe.
\newblock Reverse engineering modeling methods and tools: a survey.
\newblock {\em Computer-Aided Design and Applications}, 2018.

\bibitem{carion2020end}
Nicolas Carion, Francisco Massa, Gabriel Synnaeve, Nicolas Usunier, Alexander Kirillov, and Sergey Zagoruyko.
\newblock End-to-end object detection with transformers.
\newblock In {\em ECCV}, 2020.

\bibitem{onshapelink}
Paul Chastell.
\newblock Under the hood: Onshape sketches, 2015.
\newblock Accessed: 2024-15-07.

\bibitem{chen2023editable}
Ye Chen, Bingbing Ni, Xuanhong Chen, and Zhangli Hu.
\newblock Editable image geometric abstraction via neural primitive assembly.
\newblock In {\em ICCV}, 2023.

\bibitem{dosovitskiy2020image}
Alexey Dosovitskiy, Lucas Beyer, Alexander Kolesnikov, Dirk Weissenborn, Xiaohua Zhai, Thomas Unterthiner, Mostafa Dehghani, Matthias Minderer, Georg Heigold, Sylvain Gelly, Jakob Uszkoreit, and Neil Houlsby.
\newblock An image is worth 16x16 words: Transformers for image recognition at scale.
\newblock In {\em ICLR}, 2021.

\bibitem{dupont2022cadops}
Elona Dupont, Kseniya Cherenkova, Anis Kacem, Sk~Aziz Ali, Ilya Arzhannikov, Gleb Gusev, and Djamila Aouada.
\newblock Cadops-net: Jointly learning cad operation types and steps from boundary-representations.
\newblock In {\em 3DV}, 2022.

\bibitem{dupont2024transcad}
Elona Dupont, Kseniya Cherenkova, Dimitrios Mallis, Gleb Gusev, Anis Kacem, and Djamila Aouada.
\newblock Transcad: A hierarchical transformer for cad sequence inference from point clouds.
\newblock {\em ECCV}, 2024.

\bibitem{egiazarian2020deep}
Vage Egiazarian, Oleg Voynov, Alexey Artemov, Denis Volkhonskiy, Aleksandr Safin, Maria Taktasheva, Denis Zorin, and Evgeny Burnaev.
\newblock Deep vectorization of technical drawings.
\newblock In {\em ECCV}, 2020.

\bibitem{ganin2021computer}
Yaroslav Ganin, Sergey Bartunov, Yujia Li, Ethan Keller, and Stefano Saliceti.
\newblock Computer-aided design as language.
\newblock In {\em NeurIPS}, 2021.

\bibitem{sketchrnnICLR2018}
David Ha and Douglas Eck.
\newblock A neural representation of sketch drawings.
\newblock In {\em ICLR}, 2018.

\bibitem{he2022masked}
Kaiming He, Xinlei Chen, Saining Xie, Yanghao Li, Piotr Doll{\'a}r, and Ross Girshick.
\newblock Masked autoencoders are scalable vision learners.
\newblock In {\em CVPR}, 2022.

\bibitem{He2015DeepRL}
Kaiming He, X. Zhang, Shaoqing Ren, and Jian Sun.
\newblock Deep residual learning for image recognition.
\newblock In {\em CVPR}, 2015.

\bibitem{huang2019learning}
Zhewei Huang, Wen Heng, and Shuchang Zhou.
\newblock Learning to paint with model-based deep reinforcement learning.
\newblock In {\em ICCV}, 2019.

\bibitem{karadeniz2024davinci}
Ahmet~Serdar Karadeniz, Dimitrios Mallis, Nesryne Mejri, Kseniya Cherenkova, Anis Kacem, and Djamila Aouada.
\newblock Davinci: A single-stage architecture for constrained cad sketch inference.
\newblock In {\em BMVC}, 2024.

\bibitem{khan2024cad}
Mohammad~Sadil Khan, Elona Dupont, Sk~Aziz Ali, Kseniya Cherenkova, Anis Kacem, and Djamila Aouada.
\newblock Cad-signet: Cad language inference from point clouds using layer-wise sketch instance guided attention.
\newblock In {\em CVPR}, 2024.

\bibitem{Kingma2014AdamAM}
Diederik~P. Kingma and Jimmy Ba.
\newblock Adam: A method for stochastic optimization.
\newblock In {\em ICLR}, 2015.

\bibitem{kuhn1955hungarian}
Harold~W Kuhn.
\newblock The hungarian method for the assignment problem.
\newblock {\em Naval research logistics quarterly}, 1955.

\bibitem{li2020differentiable}
Tzu-Mao Li, Michal Luk{\'a}{\v{c}}, Micha{\"e}l Gharbi, and Jonathan Ragan-Kelley.
\newblock Differentiable vector graphics rasterization for editing and learning.
\newblock {\em ACM TOG}, 2020.

\bibitem{li2015free}
Yi Li, Timothy~M Hospedales, Yi-Zhe Song, and Shaogang Gong.
\newblock Free-hand sketch recognition by multi-kernel feature learning.
\newblock {\em CVIU}, 2015.

\bibitem{ma2022towards}
Xu Ma, Yuqian Zhou, Xingqian Xu, Bin Sun, Valerii Filev, Nikita Orlov, Yun Fu, and Humphrey Shi.
\newblock Towards layer-wise image vectorization.
\newblock In {\em CVPR}, 2022.

\bibitem{mallis2023sharp}
Dimitrios Mallis, Ali~Sk Aziz, Elona Dupont, Kseniya Cherenkova, Ahmet~Serdar Karadeniz, Mohammad~Sadil Khan, Anis Kacem, Gleb Gusev, and Djamila Aouada.
\newblock Sharp challenge 2023: Solving cad history and parameters recovery from point clouds and 3d scans. overview, datasets, metrics, and baselines.
\newblock In {\em ICCVW}, 2023.

\bibitem{mo2021general}
Haoran Mo, Edgar Simo-Serra, Chengying Gao, Changqing Zou, and Ruomei Wang.
\newblock General virtual sketching framework for vector line art.
\newblock {\em ACM TOG}, 2021.

\bibitem{Nakano2019NeuralPA}
Reiichiro Nakano.
\newblock Neural painters: A learned differentiable constraint for generating brushstroke paintings.
\newblock {\em ArXiv}, 2019.

\bibitem{para2021sketchgen}
Wamiq Para, Shariq Bhat, Paul Guerrero, Tom Kelly, Niloy Mitra, Leonidas~J Guibas, and Peter Wonka.
\newblock Sketchgen: Generating constrained cad sketches.
\newblock In {\em NeurIPS}, 2021.

\bibitem{Parmar2018ImageT}
Niki Parmar, Ashish Vaswani, Jakob Uszkoreit, Lukasz Kaiser, Noam~M. Shazeer, Alexander Ku, and Dustin Tran.
\newblock Image transformer.
\newblock In {\em ICML}, 2018.

\bibitem{Paszke2017AutomaticDI}
Adam Paszke, Sam Gross, Soumith Chintala, Gregory Chanan, Edward Yang, Zach DeVito, Zeming Lin, Alban Desmaison, Luca Antiga, and Adam Lerer.
\newblock Automatic differentiation in pytorch.
\newblock 2017.

\bibitem{qi2022one}
Anran Qi, Yulia Gryaditskaya, Tao Xiang, and Yi-Zhe Song.
\newblock One sketch for all: One-shot personalized sketch segmentation.
\newblock {\em IEEE TIP}, 2022.

\bibitem{reddy2021im2vec}
Pradyumna Reddy, Michael Gharbi, Michal Lukac, and Niloy~J Mitra.
\newblock Im2vec: Synthesizing vector graphics without vector supervision.
\newblock In {\em CVPR}, 2021.

\bibitem{Reddy2021Im2VecSV}
Pradyumna Reddy, Micha{\"e}l Gharbi, Michal Luk{\'a}c, and Niloy~Jyoti Mitra.
\newblock Im2vec: Synthesizing vector graphics without vector supervision.
\newblock In {\em CVPR}, pages 7338--7347, 2021.

\bibitem{Ronneberger2015UNetCN}
Olaf Ronneberger, Philipp Fischer, and Thomas Brox.
\newblock U-net: Convolutional networks for biomedical image segmentation.
\newblock In {\em MICCAI}. Springer, 2015.

\bibitem{schmidt2019generalization}
Florian Schmidt.
\newblock Generalization in generation: A closer look at exposure bias.
\newblock {\em arXiv preprint arXiv:1910.00292}, 2019.

\bibitem{seff2020sketchgraphs}
Ari Seff, Yaniv Ovadia, Wenda Zhou, and Ryan~P. Adams.
\newblock Sketch{G}raphs: A large-scale dataset for modeling relational geometry in computer-aided design.
\newblock In {\em ICMLW}, 2020.

\bibitem{seff2022vitruvion}
Ari Seff, Wenda Zhou, Nick Richardson, and Ryan~P Adams.
\newblock Vitruvion: A generative model of parametric cad sketches.
\newblock In {\em ICLR}, 2022.

\bibitem{vaswani2017attention}
Ashish Vaswani, Noam Shazeer, Niki Parmar, Jakob Uszkoreit, Llion Jones, Aidan~N Gomez, {\L}ukasz Kaiser, and Illia Polosukhin.
\newblock Attention is all you need.
\newblock In {\em NeurIPS}, 2017.

\bibitem{wang2024parametric}
Xiaogang Wang, Liang Wang, Hongyu Wu, Guoqiang Xiao, and Kai Xu.
\newblock Parametric primitive analysis of cad sketches with vision transformer.
\newblock {\em IEEE Transactions on Industrial Informatics}, 2024.

\bibitem{willis2021engineering}
Karl~DD Willis, Pradeep~Kumar Jayaraman, Joseph~G Lambourne, Hang Chu, and Yewen Pu.
\newblock Engineering sketch generation for computer-aided design.
\newblock In {\em CVPR}, 2021.

\bibitem{wu2021density}
Tong Wu, Liang Pan, Junzhe Zhang, Tai Wang, Ziwei Liu, and Dahua Lin.
\newblock Density-aware chamfer distance as a comprehensive metric for point cloud completion.
\newblock {\em arXiv preprint arXiv:2111.12702}, 2021.

\bibitem{xu2022deep}
Peng Xu, Timothy~M Hospedales, Qiyue Yin, Yi-Zhe Song, Tao Xiang, and Liang Wang.
\newblock Deep learning for free-hand sketch: A survey.
\newblock {\em IEEE TPAMI}, 2022.

\bibitem{xu2021inferring}
Xianghao Xu, Wenzhe Peng, Chin-Yi Cheng, Karl~DD Willis, and Daniel Ritchie.
\newblock Inferring cad modeling sequences using zone graphs.
\newblock In {\em CVPR}, 2021.

\bibitem{yang2021sketchgnn}
Lumin Yang, Jiajie Zhuang, Hongbo Fu, Xiangzhi Wei, Kun Zhou, and Youyi Zheng.
\newblock Sketchgnn: Semantic sketch segmentation with graph neural networks.
\newblock {\em ACM TOG}, 2021.

\bibitem{yang2022discovering}
Yuezhi Yang and Hao Pan.
\newblock Discovering design concepts for cad sketches.
\newblock In {\em NeurIPS}, 2022.

\bibitem{yu2017sketch}
Qian Yu, Yongxin Yang, Feng Liu, Yi-Zhe Song, Tao Xiang, and Timothy~M Hospedales.
\newblock Sketch-a-net: A deep neural network that beats humans.
\newblock {\em IJCV}, 2017.

\bibitem{Zheng2018StrokeNetAN}
N. Zheng, Yifan Jiang, and Ding Huang.
\newblock Strokenet: A neural painting environment.
\newblock In {\em ICLR}, 2018.

\end{thebibliography}
}

\clearpage

\maketitlesupplementary

\maketitle

\begin{strip}\centering
    \includegraphics[width=\linewidth]{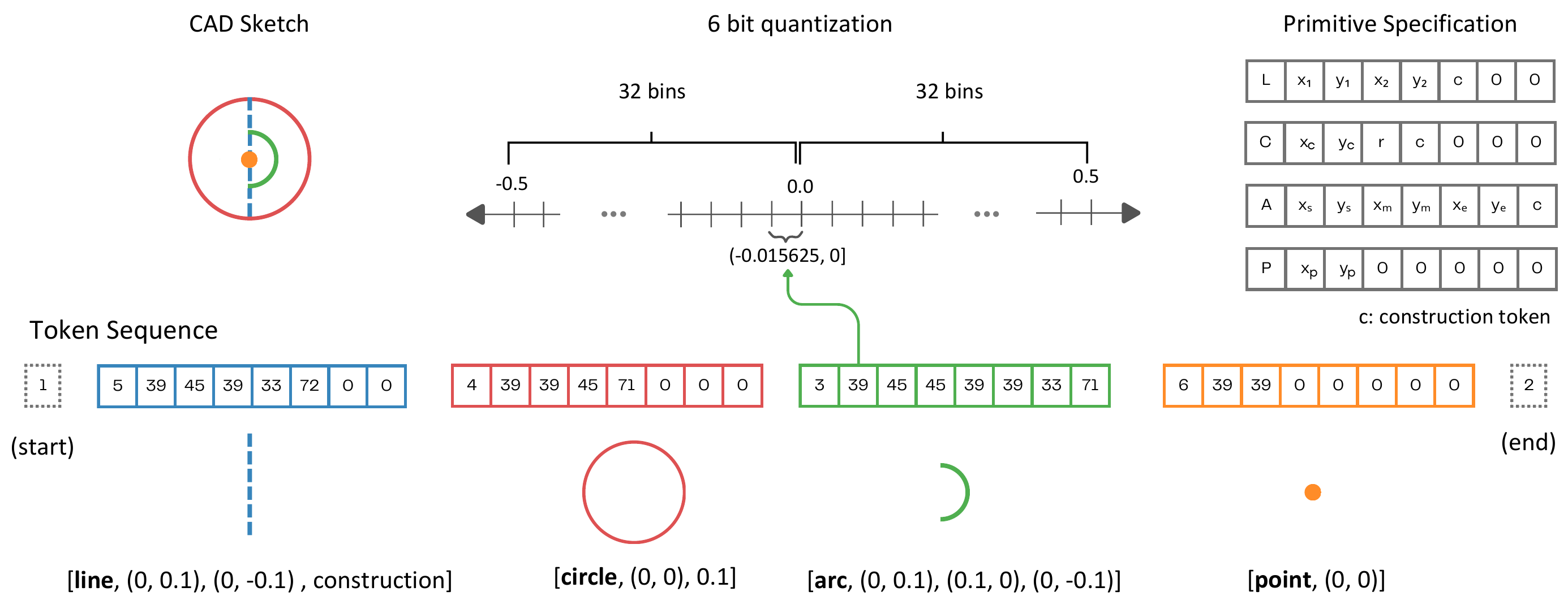}
    \captionof{figure}{Overview of the tokenization process. A parameterized CAD sketch is represented by a set of primitives, each comprising a sequence of 8 tokens. The first primitive token specifies the primitive type, followed by quantized primitive parameters. As depicted in the example, the parametric value $0.0$ is mapped on the bin $(-0.015625,0]$ due to the 6-bit quantization. Additional tokens are padded with the $0$ token value. The primitive sequence is initiated with the \textit{start} token and completed by the \textit{end} token. }
    \label{fig:tokenization}
\end{strip}

This supplementary material includes various details that were not reported in the main paper due to space constraints. To demonstrate the benefit of the proposed \texttt{PICASSO}, we also expand our experimental evaluation.

\section{System Details}
We start by reiterating details of the tokenization strategy followed by \texttt{PICASSO}. This section also discusses the effect of the multiscale loss $\mathcal{L}_{ml2}$ and reports inference times for our proposed method and that of~\cite{seff2022vitruvion}.

\subsection{Tokenization}
For our problem formulation, each primitive $\mathbf{p}_i$ is expressed as a collection of $8$ tokens $\{t_i^j\}_{j \in [1,8]}$ with $t_i^j \in [0,72]$ that capture both types and quantized primitive parameters. A detailed description of token types and corresponding token values is shown in Table \ref{tab:tokentypes}. In Fig.~\ref{fig:tokenization}, we present an overview of the tokenization process.

\begin{table}[t]
    \centering
    \begin{tabular}{l|l}
         \textit{Token Value} & \textit{Token Description} \\
         \toprule
         0 & Padding\\
         1 & Start\\
         2 & End\\
         3 & Arc\\
         4 & Circle\\
         5 & Line\\
         6 & Point\\
         $[7,70]$ & Quantized primitive parameters\\
         71 & Construction Primitive\\
         72 & Non-Construction Primitive\\
         \bottomrule
         
    \end{tabular}
    \caption{Description of tokens used in our problem formulation. }
    \label{tab:tokentypes}
\end{table}

\subsection{Multiscale $l2$ loss.}

\begin{figure}[t]
    \centering
    \includegraphics[width=\linewidth]{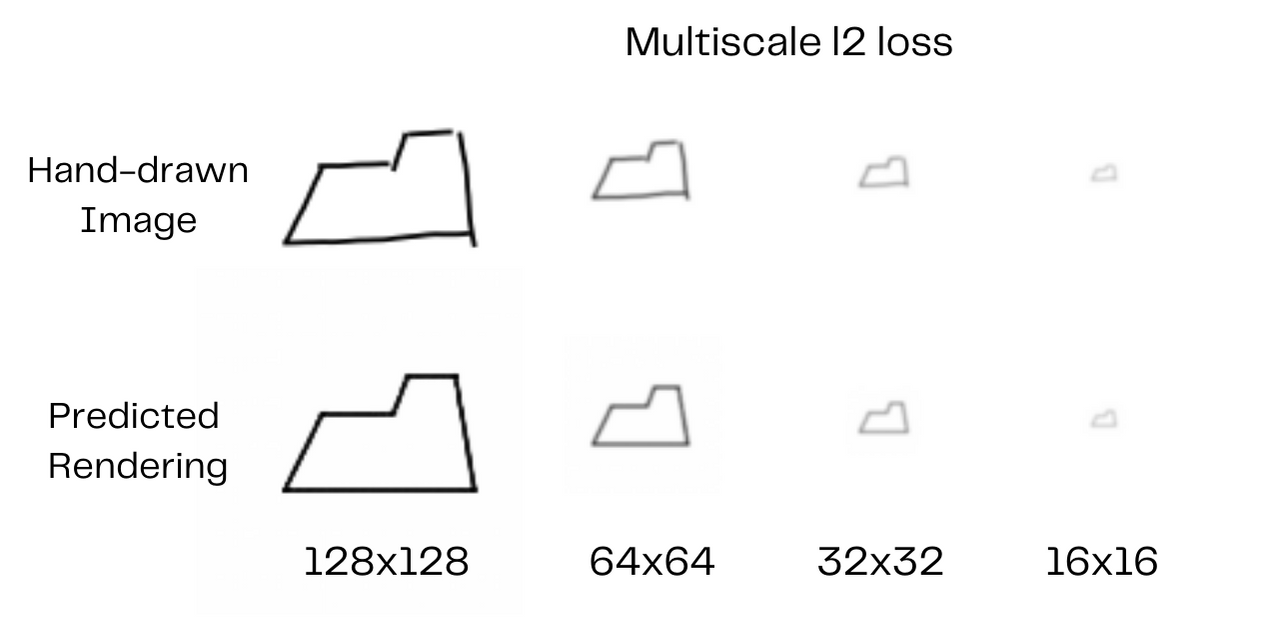}
    \caption{Illustration of the $\mathcal{L}_{ml2}$ for visually supervised CAD parameterization. The immediate support between the predicted rendering and the imprecise hand-drawn sketch image increases at lower resolutions. This mechanism compensates for noisy gradient estimates due to partial overlap at higher resolutions.}
    \label{fig:l2loss}
\end{figure}

\begin{figure}[h]
    \centering
	\includegraphics[width=\linewidth]{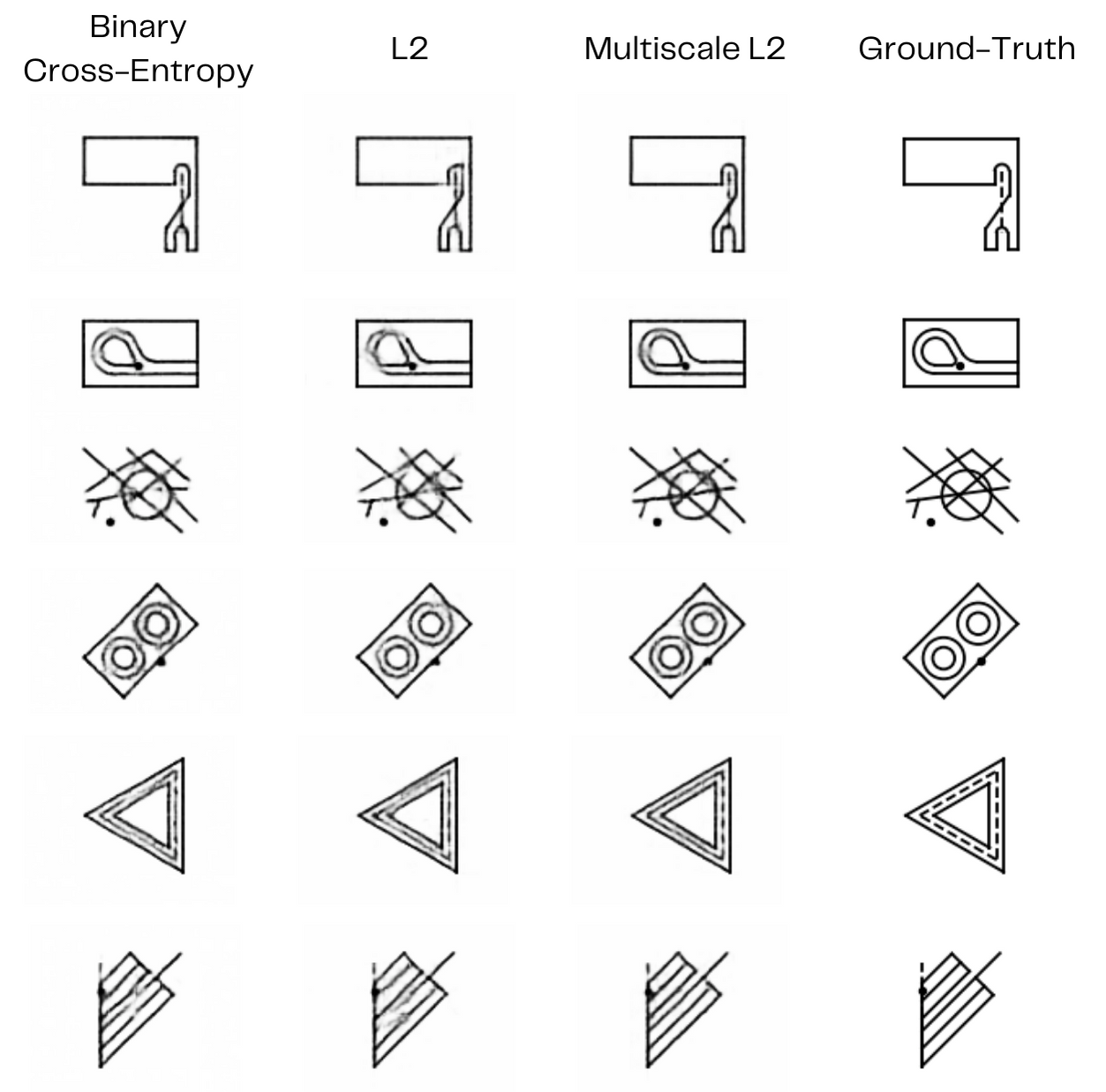}
    \caption{Sketch renderings by the SRN-\texttt{PICASSO}, trained using different image-level losses. Training SNR via a multiscale $l2$ loss results in sharp CAD sketch renderings. In contrast, employing binary cross-entropy or a standard $l2$
  loss during the learning process may lead to renderings that are blurred, lack fine details, or have disconnected primitives. }
    \label{fig:l2loss_rendering}
\end{figure}

Rendering self-supervision via the proposed Sketch Rendering Network (SRN) is facilitated by a multiscale $l2$ loss denoted by $\mathcal{L}_{ml2}$. The multiscale $l2$ loss enables effective rendering self-supervision for precise as well as hand-drawn sketch images. Even though discrepancies are inevitably introduced due to the imprecise nature of hand-drawn lines, the loss at a lower resolution can still provide an informative learning signal. A visualisation of image pyramids constructed for $\mathcal{L}_{ml2}$ computation is presented in Fig.~\ref{fig:l2loss}. Qualitative results of renderings produced by SRN when trained with different image-level losses are given in Fig.~\ref{fig:l2loss_rendering}. We observe that utilizing the multiscale $l2$ loss during training, results in sharper images and accurate rendering of finer details. Improved SRN renderings in turn lead to the computation of informative gradients that enable rendering self-supervision and zero / few-shot learning scenarios for  \texttt{PICASSO}, as demonstrated in Sections 5.2 of the main paper.

\subsection{Inference Time Comparison}
In Table~\ref{tab:timenalaysis}, we report inference time in seconds for our method and that of \cite{seff2022vitruvion}. We observe that our Sketch Parametrization Network (SPN) enables faster inference. In contrast, Vitruvion~\cite{seff2022vitruvion} is an autoregressive method that requires multiple forward passes per sample, leading to increased total inference time.

\begin{table}[h]
    \centering
    \begin{tabular}{lc}
       Method & Inference~(\textit{sec}) \\
        \toprule
        Vitruvion~\cite{seff2022vitruvion} & 1.1005\\
        SPN-\texttt{PICSASSO} & 0.1101\\
        \bottomrule
    \end{tabular}
    \caption{Inference times per-sample for our proposed method and that of~\cite{seff2022vitruvion}. Results are computed for a batch size of 1.}
    \label{tab:timenalaysis}
\end{table}

\section{Metrics}
The metrics utilized for quantitative evaluation are detailed as follows.

\vspace{0.2cm}
\noindent \textbf{Acc:}
To enable the computation of parameter-based metrics, the permutation $\hat{\pi} \in \Pi_n$ of the predicted w.r.t the ground truth is recovered with bipartite matching. Accuracy is computed as:

\begin{equation}
\textit{Acc} = \frac{1}{T_{Acc}}\sum_{i=1}^{n_z}\sum_{j=1}^{8} \mathbbm{1}[t_i^j > 0] \cdot \mathbbm{1}[t_i^j = \hat{t}_{\hat{\pi}(i)}^j]
\end{equation}

\noindent
where $\hat{t}_{\hat{\pi}(i)}^j$ and $t^j_i$ and the predicted and ground truth token sequences respectively, $n_z = |\{\boldsymbol{p}^z\}|$ is the number of primitives for the $z$'th test sample, $T_{Acc}$ is the number of non-padding tokens in the ground truth sequence or formally $T_{Acc} = \sum_{i=1}^{n_z}\sum_{j=1}^{8} \mathbbm{1}[t_i^j > 0]$,  and $\mathbbm{1}[.]$ is the indicator function.

\vspace{0.2cm}
\noindent \textbf{ParamMSE:}
Parametric \textit{mean-squared error (ParamMSE)} considers solely the parameter tokens of predicted primitives, thus type, padding, and construction tokens are excluded. Formally,

\begin{equation}
\text{\small \textit{ParamMSE}} = \frac{1}{T_{MSE}}\sum_{i=1}^{n_z}\sum_{j=1}^{8} \mathbbm{1}[t_i^j > 6] \cdot \mathbbm{1}[t_i^j < 71] \cdot (t_i^j - \hat{t}_{\hat{\pi}(i)}^j)^2
\end{equation}

\noindent
where $T_{MSE} = \sum_{i=1}^{n_z}\sum_{j=1}^{8} \mathbbm{1}[t_i^j > 6] \cdot \mathbbm{1}[t_i^j < 71]$
 is the number of parameter tokens in the ground truth sequence.

\vspace{0.2cm}
\noindent \textbf{ImgMSE:}
The pixel-wise \textit{Mean Squared Error} \textit{ImgMSE} comprise two MSE terms. Formally:

\begin{equation}
\begin{split}
\textit{ImgMSE} =  \frac{1}{2N_F}
\sum_{k=1}^{w \cdot h} \mathbbm{1}[ \mathbf{X}_k = 1] \cdot
(\mathbf{\hat{X}}_k - \mathbf{X}_k)^2 \\
+ \frac{1}{2w \cdot h}
\sum_{k=1}^{w \cdot h} 
(\mathbf{\hat{X}}_k - \mathbf{X}_k)^2 
\end{split}
\end{equation}

\noindent
where $N_F=\sum_{k=1}^{w \cdot h} \mathbbm{1}[ \mathbf{X}_k = 1]$ is the number of foreground  pixels.

\vspace{0.2cm}
\noindent \textbf{CD:}
To compute bidirectional Chamfer Distance (CD), we form a set of foreground pixel coordinates $\boldsymbol{\zeta} \in \{(i,j) \mid 1 \leq i \leq h, \, 1 \leq j \leq w\}$ for both the ground truth and predicted explicit renderings. The result is the sets $\mathrm{Z} =\{{\boldsymbol{\zeta}}_n\}_{n=1}^{N_f}$ and $\mathrm{\hat{Z}} =\{{\boldsymbol{\hat{\zeta}}}_n\}_{n=1}^{\hat{N}_f}$ where $N_f$ and $\hat{N}_f$ is the number of foreground pixels for ground truth and prediction explicit renderings respectively. Bi-directional chamfer distance is given by:

\begin{equation}
CD = \frac{1}{2\hat{N}_{f}} \sum_{n=1}^{\hat{N}_{f}} \min _{\boldsymbol{\zeta}_k \in \mathrm{Z}}\|\hat{{\boldsymbol{\zeta}}}_n-\boldsymbol{\zeta}_k\|_2^2 \ 
+ \frac{1}{{2N}_{f}} \sum_{n=1}^{N_{f}} \min _{\hat{\boldsymbol{\zeta}_k} \in \hat{\mathrm{Z}}}\|{{\boldsymbol{\zeta}}}_n-\hat{\boldsymbol{\zeta}_k}\|_2^2 ,
    \label{eq:CD1}
\end{equation}

\section{Implementation Details for Comparative Analysis}

To evaluate the effectiveness of the proposed \texttt{PICASSO}, comparisons are performed in two settings; \textit{(1)} For few-shot \wrt the state-of-the-art autoregressive method of Vitruvion~\cite{seff2022vitruvion} and a non-autoregressive baseline based on a convolutional ResNet34 backbone and \textit{(2)} for zero-shot w.r.t the Primitive Matching Network of~\cite{alaniz2022abstracting}. The proposed SRN is also contrasted to the differentiable renderer DiffVG~\cite{li2020differentiable}. This section will expand on implementation details related to these methods.

\vspace{0.2cm}
\noindent
\textbf{Vitruvion:} We train Vitruvion~\cite{seff2022vitruvion} on the SketchGraphs~\cite{seff2020sketchgraphs} dataset using the publicly available implementation\footnote{\url{https://github.com/PrincetonLIPS/vitruvion}}. For autoregressive CAD parameterization, we select the next token via $argmax$ instead of the nucleus sampling used for sketch generation. This modification enhances parameterization performance, while also ensuring consistent reproducibility of results. All hyperparameters are set as in the original paper~\cite{seff2022vitruvion}.

\vspace{0.2cm}
\noindent \textbf{ResNet34:} To form a non-autoregressive baseline we trained a ResNet34 followed by global pooling. The output of the convolutional backbone is fed into a Multi-Layer Perceptron (\textit{MLP}) with 2 linear layers and a ReLU activation. The final token predictions are produced by a softmax on the output logits of the MLP.

\vspace{0.2cm}
\noindent \textbf{PpaCAD:} We implemented the concurrent method PpaCAD~\cite{wang2024parametric} as no public code is available. Image encoding uses a 2-layer patch MLP followed by a transformer. Separate losses are computed for each primitive type, parameter, and construction flag. Further details are in ~\cite{wang2024parametric}.

\vspace{0.2cm}
\noindent \textbf{Primitive Matching Network (PMN):} For comparison with PMN, we re-train on the Sketchgraphs dataset using the publicly available code\footnote{\url{https://github.com/ExplainableML/sketch-primitives}}. We form strokes by sampling points on parametric primitives that are provided as input to PMN. The input for PMN comprises multiple sets of coordinates, with each set uniquely representing a single distinct primitive. 
Note that this scenario presents a less complex challenge than that encountered in \texttt{PICASSO}, where parameterization is derived directly from raster images. In such case, primitives may overlap or be positioned in close proximity, significantly increasing the complexity of the parameterization task. Strokes are processed by PMN to obtain \textit{drawing primitives} that consist of the abstracted output stroke and the stroke type (line, circle, half-circle and point). Finally, the output strokes are parameterized via least-square fitting based on their predicted types.

\vspace{0.2cm}
\noindent
\textbf{DiffVG:} Comparison to DiffVG~\cite{li2020differentiable} is performed on 1) pre-training and 2) test-time optimization settings as discussed in Section 5.3 of the main paper. Predicted sequences are transformed into Bezier paths to enable an image-level loss. For the pre-training setting, SPN is trained with respect to the image loss. For test-time optimization, the paths are iteratively updated through differentiable rendering. As already noted, DiffVG can only update path parameters, but it is unable to change discrete decisions like path types. Lines are converted to paths with two endpoints and no control points. Points are also composed of 2 endpoints formed by shifting the point coordinate by 1 quantization unit. Arcs and circles are formed by Bezier paths, computed via the Python package~\href{https://github.com/mathandy/svgpathtools}{svgpathtools}\footnote{\url{https://github.com/mathandy/svgpathtools}}. After optimization, paths are converted back to the considered primitives (lines, arcs, circles, and points) and evaluated directly w.r.t ground truth sequences.

\section{Semi-Supervised CAD Sketch Parameterization via Rendering Supervision}

\begin{table}
    \centering
        \setlength{\belowcaptionskip}{-0.5cm}
    \setlength{\tabcolsep}{4pt}
    \resizebox{\linewidth}{!}{
    \begin{tabular}{lcccc}
        \toprule
        Method & \textit{Acc} & \textit{ParamMSE} & \textit{ImgMSE} & \textit{CD} \\
        \midrule
        PICASSO (w/o pt.) & 0.595 & 451 & 0.156 & 2.789 \\
        PICASSO (w/o pt. + semi-supervised) &   0.608 & 432 & 0.145 & 1.833  \\
        \bottomrule
    \end{tabular}
    }
    \vspace{-0.1cm}
    \caption{Semi-supervised learning results for \texttt{PICASSO}.}
    \label{tab:semisupervised}
\end{table}

Rendering supervision enabled by the proposed SRN can also be applied to other learning schemes. We investigate a semi-supervised learning scenario where SRN is trained through rendering supervision on unlabelled sketch images and parametric supervision on a smaller set of parameterized sketches (16k samples). Table~\ref{tab:semisupervised} shows quantitative results of the semi-supervised \texttt{PICASSO} compared to its parametrically supervised counterpart on 16k samples. By leveraging unlabelled sketches through rendering supervision, the semi-supervised model can achieve better performance. The difference is more noticeable in terms of image-based metrics, as rendering supervision can result in a model that discovers plausible geometric reconstructions, that might depart from the ground truth CAD sketch parameterization.

\section{Sensitivity to Rendering Quality}

\begin{figure}[h]
    \centering
	\includegraphics[width=0.65\linewidth]{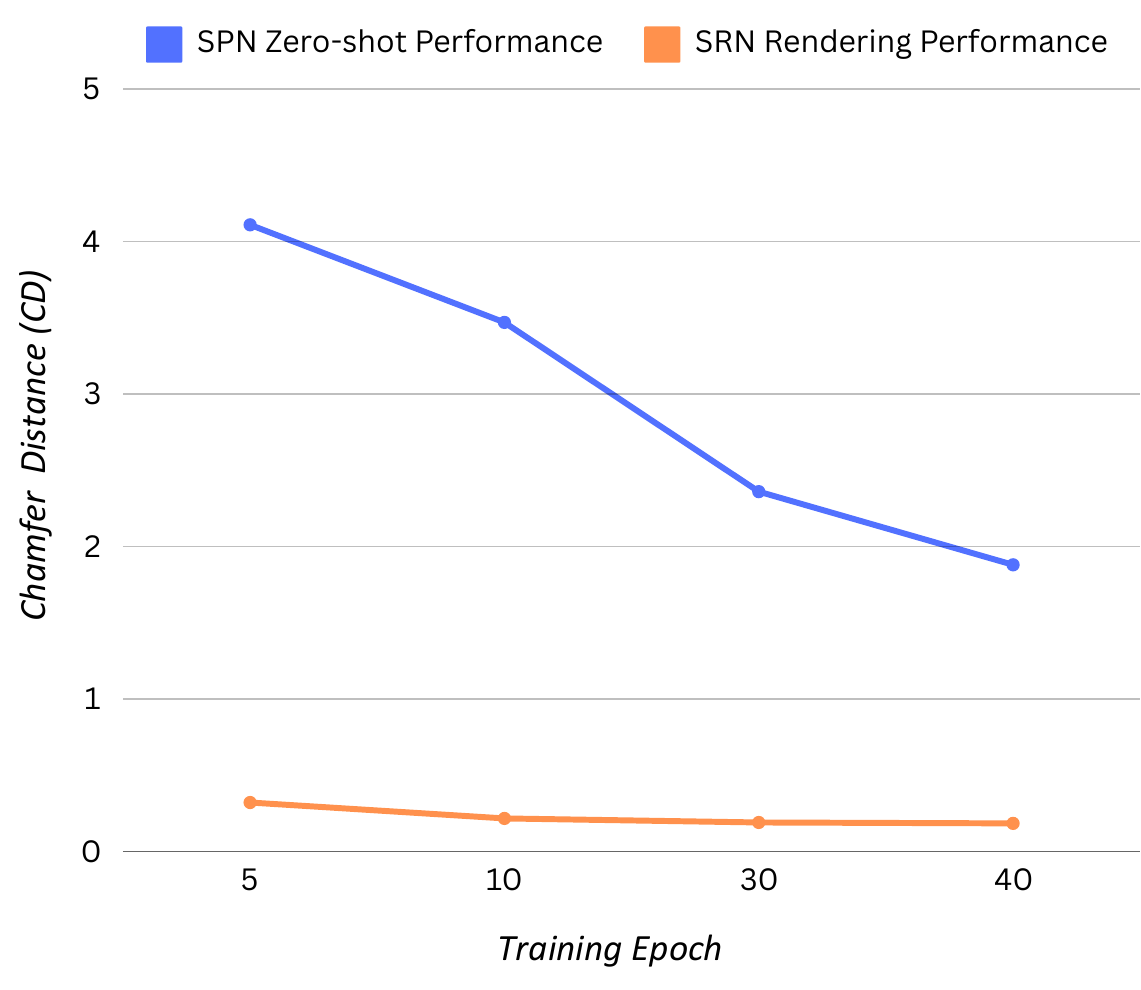}
    \vspace{-0.3cm}
    \caption{\textit{(orange)} Rendering performance (in terms of chamfer distance) of SRN trained for different number of epochs. \textit{(blue)} Zero-shot performance of \texttt{PICASSO}-SPN when pretrained by the aforementioned SRN renderers.}
    \label{fig:zeroshot_robustness}
\end{figure}

For \texttt{PICASSO}, rendering self-supervision is facilitated by neural differentiable rendering via SRN. We conduct an ablation study to explore how variations in SRN's rendering performance influence the effectiveness of self-supervised pretraining. To that end, we vary SRN rendering quality by training SRN for different number of epochs. Specifically, \texttt{PICASSO} is pretrained via rendering self-supervision using different SRN renderers trained for 5, 10, 30 and 40 epochs.  The results are presented in Figure~\ref{fig:zeroshot_robustness}. We find that SRN rendering performance (measured in terms of chamfer distance) stabilizes after a few training epochs \textit{(orange line)}.  However, the zero-shot performance of SPN is notably stronger when using a fully trained SRN \textit{(blue line)}. As the neural rendering improves, fewer artifacts are introduced and SRN can more accurately replicates the input sketch parameters.

\section{Extension to Other Primitives}

\begin{figure}[h]
\setlength{\belowcaptionskip}{-0.5cm}
    \centering
\includegraphics[ width=0.7\linewidth]{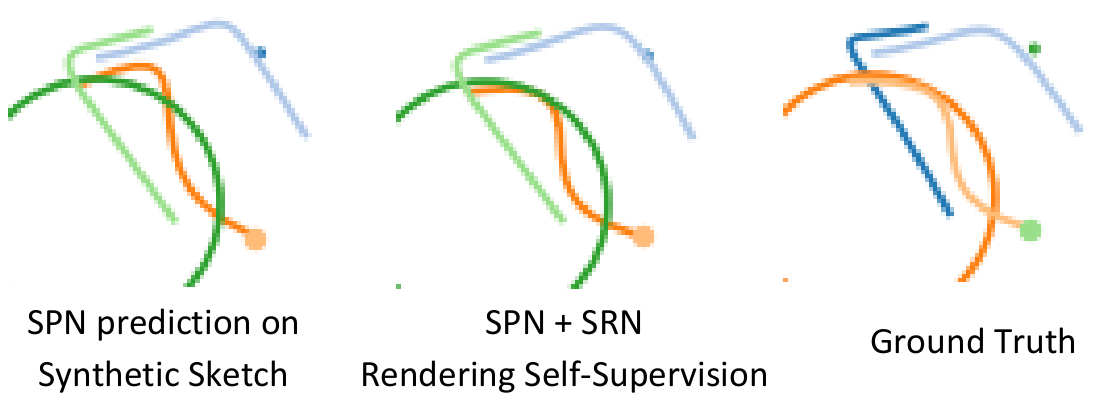}
    \caption{PICASSO on synthetic CAD sketch including B-spline.}
    \label{fig:splines}
\end{figure}

The main experiments of \texttt{PICASSO} are conducted for parameterizing lines, circles, arcs, and points. Free-form curves such as B-splines, hyperbolas, and NURBS are excluded similarly to recent works as they are underrepresented in existing datasets  (\eg b-splines are $2.57\%$ of SketchGraphs primitives~\cite{seff2020sketchgraphs}). As a preliminary experiment for future work, we trained and tested SPN and SRN on a \textit{synthetic} dataset including randomly generated b-splines. Training is performed for 20 epochs and a different random sketch is sampled at each iteration. Table~\ref{tab:splines} reports a comparison to DiffVG in terms of test time optimization on 100 synthetically generated images. SRN self-supervision improves b-spline predictions of SPN and significantly surpasses DiffVG in terms of Chamfer Distance (CD). Figure~\ref{fig:splines} shows an example where SPN prediction on a synthetic sketch with B-spline is being improved with SRN supervision.

\begin{table}[h]
    \centering
    \setlength{\tabcolsep}{4pt}
    \resizebox{0.35\linewidth}{!}{
    \begin{tabular}{lc}
        \toprule
        Method & \textit{CD} \\
        \midrule
        SPN & 1.07 \\
        SPN+DiffVG &  3.60  \\
        SPN+SRN & 0.48 \\
        \bottomrule
    \end{tabular}
    }
    \vspace{0.1cm}
    \caption{Test time optimization of sketches that include B-splines. Optimization via SRN performs better in terms of Chamfer Distance (CD).}
    \label{tab:splines}
\end{table}

\clearpage

\onecolumn
\section{Additional Qualitative Results}
This section expands on the qualitative evaluation reported in the main paper.

\subsection{Few-shot CAD Sketch Parameterization}

This subsection expands the qualitative evaluation shown in subsection 5.2 \textit{(Few-shot Evaluation)} of the main paper. Visual results for finetuning with 2k, 16k and 32k samples are shown in Fig.~\ref{fig:fewshot}. We observe that the 32k-shot setting results in robust CAD parameterization from challenging precise and hand-drawn sketch images, even though the network is trained only with a fraction of the original dataset~($\approx 2\%$ of the SketchGraph dataset~\cite{seff2020sketchgraphs}).

\begin{figure*}[h]
\setlength{\belowcaptionskip}{-0.1cm}
    \centering
\includegraphics[width=1\linewidth]{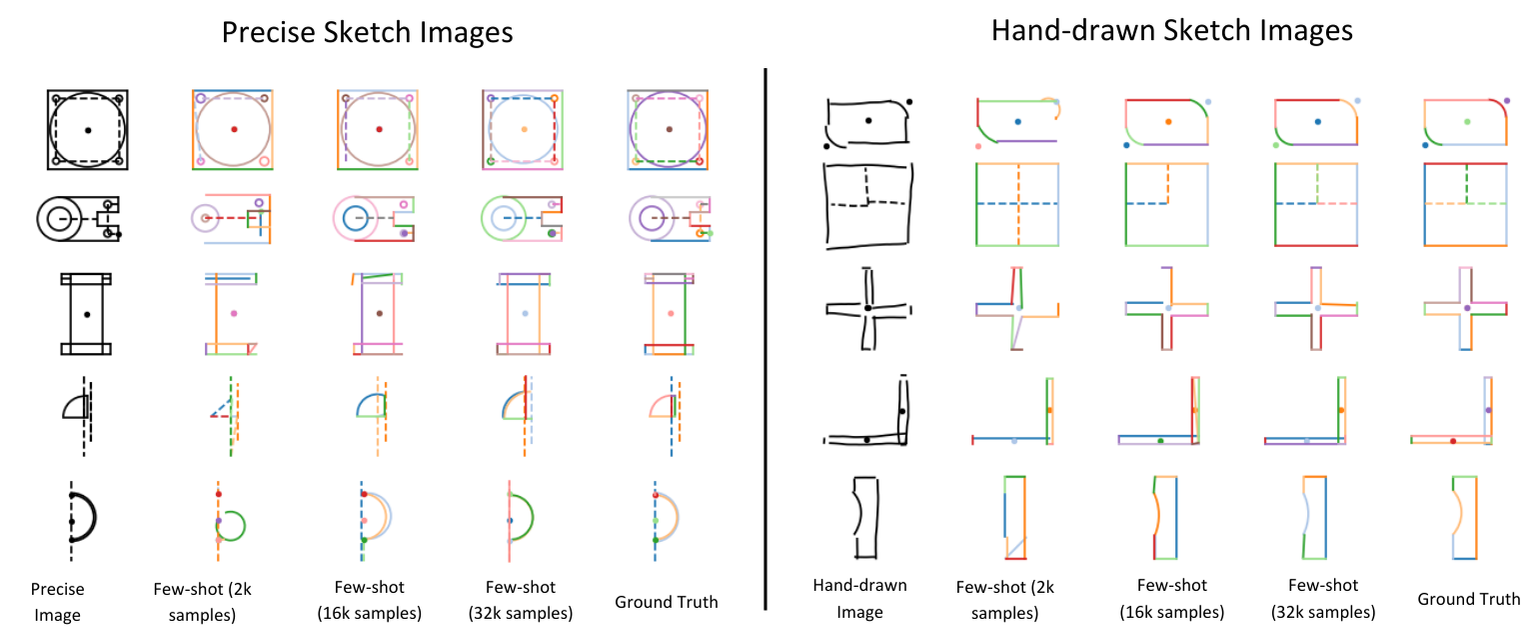}
    \caption{Few-shot setting. Qualitative results of \texttt{PICASSO} learned CAD sketch parameterization from precise and hand-drawn sketches. Best viewed in colors.}
    \label{fig:fewshot}
\end{figure*}

\begin{figure*}[h]
    \centering
	\includegraphics[width=\linewidth]{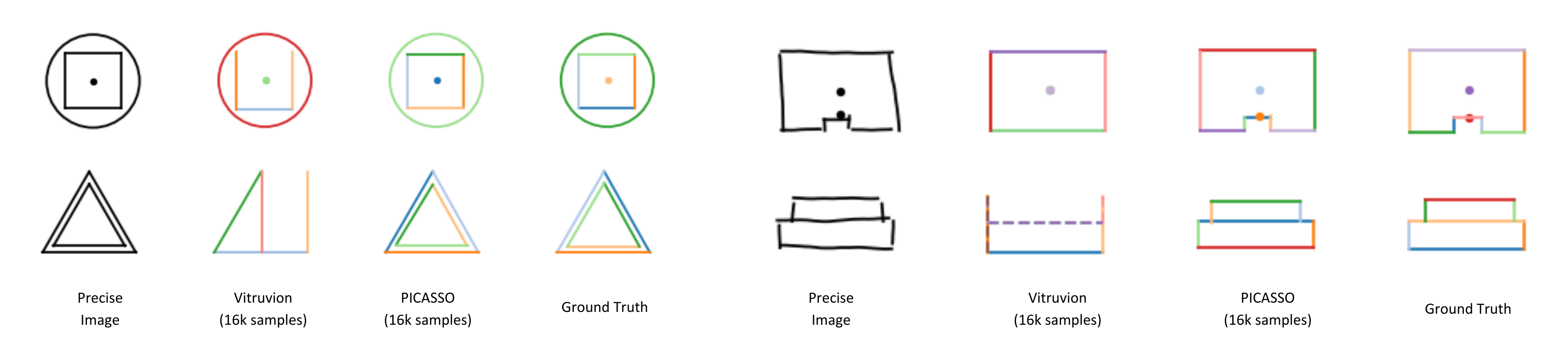}
    \caption{Visual examples of CAD sketch parameterization from hand-drawn and precise sketches by Vitruvion~\cite{seff2022vitruvion} and \texttt{PICASSO} on a 16k-shot setting.}
    \label{fig:comparison_vitr}
\end{figure*}

Fig.~\ref{fig:comparison_vitr} depicts the qualitative comparison of our model with that of Vitruvion~\cite{seff2022vitruvion} for the parameterization of precise and hand-drawn sketches. It can be observed that the proposed method produces plausible parameterizations closer to the ground truth.

\clearpage

\subsection{Zero-shot CAD Sketch Parameterization}

\begin{figure*}[h]
    \centering
	\includegraphics[width=\linewidth]{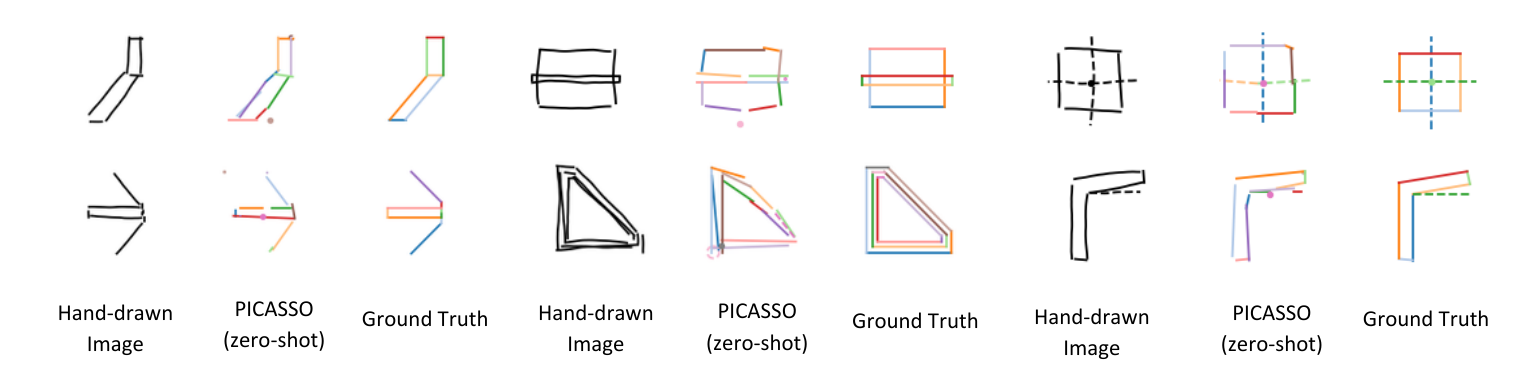}
    \caption{Qualitative results for CAD sketch parameterization of hand-drawn sketches, learned solely through rendering self-supervision with SRN-\texttt{PICASSO}. Best visualised in colors.}
    \label{fig:zeroshot}
\end{figure*}

This subsection expands the qualitative evaluation shown in subsection 5.2 \textit{(Zero-shot Evaluation)} of the main paper. In Fig.~\ref{fig:zeroshot}, we present visual examples of CAD sketch parameterization from hand-drawn sketches, learned via an image-level loss only. Under a complete lack of parametric supervision, \texttt{PICASSO} is able to roughly parameterize hand-drawn sketches. Note that compared to few-shot setting, SPN is further constrained to output a fixed number of primitives per type for the zero-shot evaluation. While \texttt{PICASSO} achieves plausible zero-shot parameterizations, we find that rendering self-supervision can be hindered by the discrepancy between hand-drawn sketches and the precise ones rendered by SRN. The development of hand-drawn invariant losses that can enhance zero-shot performance is identified as interesting future work.

\begin{figure*}[h]
    \centering
	\includegraphics[width=\linewidth]{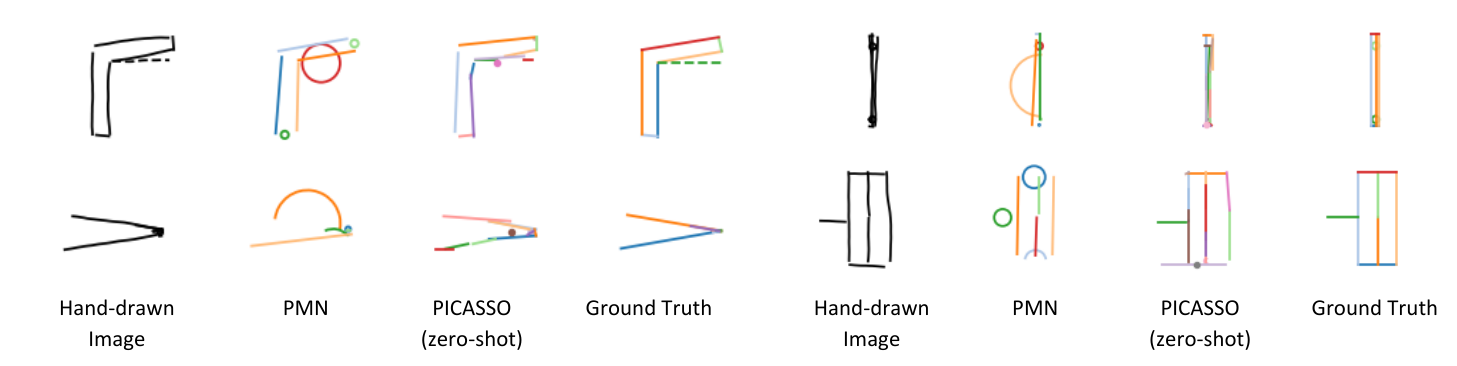}
    \caption{Zero-shot CAD sketch parameterization from hand-drawn sketches by PMN~\cite{alaniz2022abstracting} and \texttt{PICASSO}. Note that PMN has prior information on grouping individual strokes with their coordinate points, whereas PICASSO infers directly from the image space without any grouping of primitives.}
    \label{fig:comparison_pmn}
\end{figure*}

Fig.~\ref{fig:comparison_pmn} illustrates a qualitative comparison with PMN~\cite{alaniz2022abstracting} for the zero-shot setting. \texttt{PICASSO} predicts more consistent sketches with primitives that are not geometrically far from the input image. It is important to highlight that our zero-shot model works on a more challenging setup of direct parameterization from images without having access to individual groupings of strokes contrary to PMN~\cite{alaniz2022abstracting}. Also, note that we do not conduct comparison to PMN on precise images. Since PMN is aware of the grouping of distinct strokes, parameterization of precise inputs becomes a trivial task, reduced to merely identifying the types of primitive strokes.

\clearpage

\subsection{Test-time Optimization with SRN}

\begin{figure*}[h]

  \centering
\includegraphics[width=0.7\linewidth]{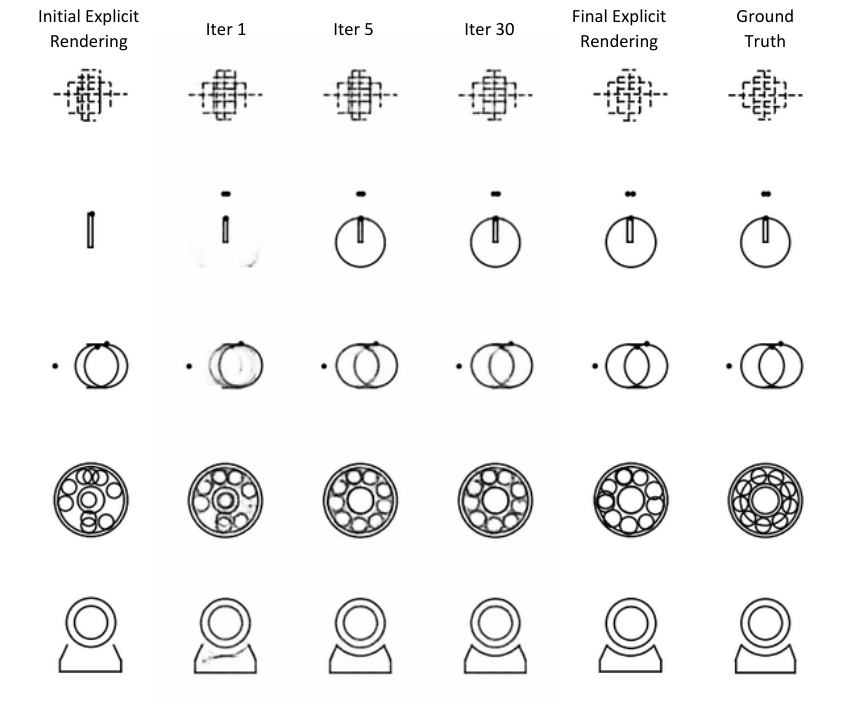}
    \caption{Test-time optimization with SRN-\texttt{PICASSO}. SRN enables the computation of an image-level loss between predicted rendering and the input precise sketch image. CAD parameterization improves over multiple backpropagation steps on a specific test sample.   }
\label{fig:test_time_sup}
\end{figure*}

As shown in subsection 5.3 \textit{(SRN vs DiffVG)} of the main paper, rendering self-supervision can be used to enhance CAD sketch parameterization produced by a parameterically supervised SPN at test-time. In Fig.~\ref{fig:test_time_sup}, we show qualitative results for test-time optimization of precise sketches. We observe that SRN can improve geometric reconstruction of CAD sketches at inference time.

\clearpage

\section{Permutation Invariance of SRN \wrt Primitive Order}

\begin{figure*}[h]
    \centering
	\includegraphics[width=0.8\linewidth]{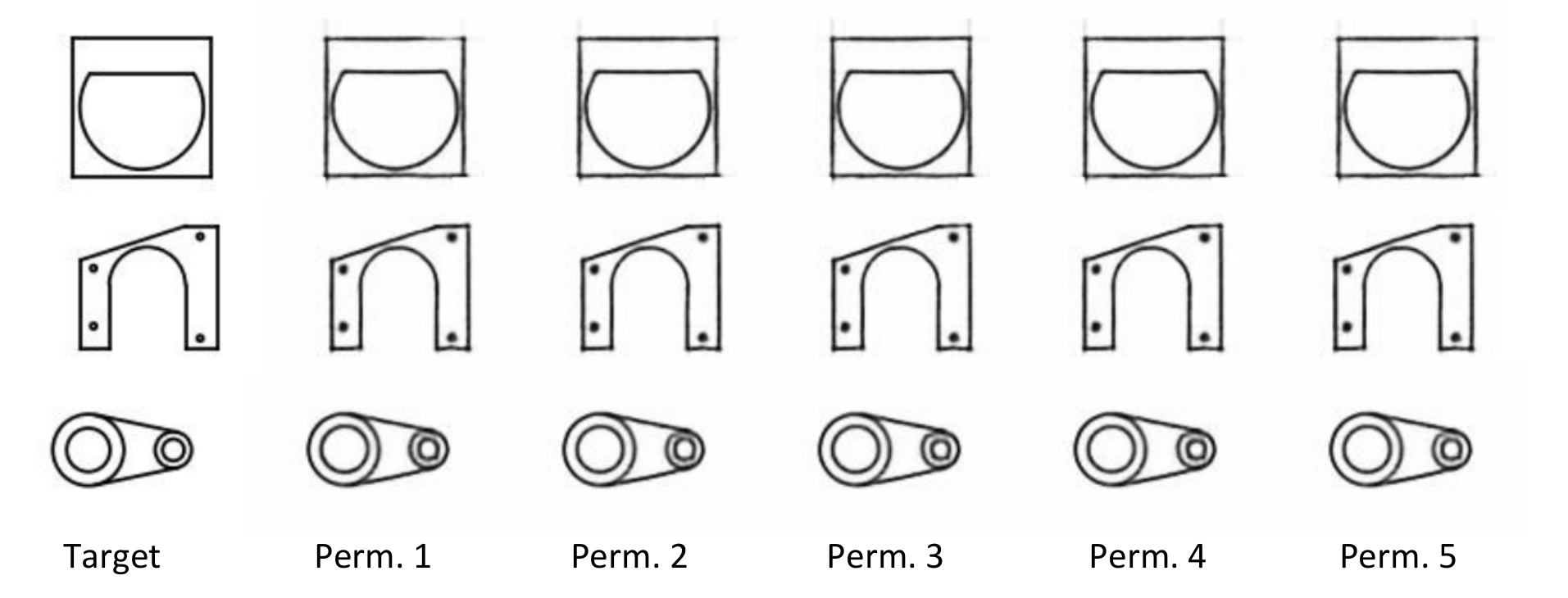}
    \caption{SRN renderings of the parametric CAD sketches where the primitives are randomly permuted.}
    \label{fig:srn_permutations}
\end{figure*}

Since SRN processes input primitive tokens as a set, it should demonstrate permutation invariance with respect to their order. Although this invariance is not explicitly designed into the architecture, it naturally emerges from the synthetic training process, where primitives can appear in any sequence position. Figure~\ref{fig:srn_permutations} qualitatively illustrates the model's robustness to input order, with renderings of CAD sketches showing only subtle differences despite permuted primitive sequences.

\section{Failure Case Analysis} 

\begin{figure*}[h]
    \centering
	\includegraphics[width=0.8\linewidth]{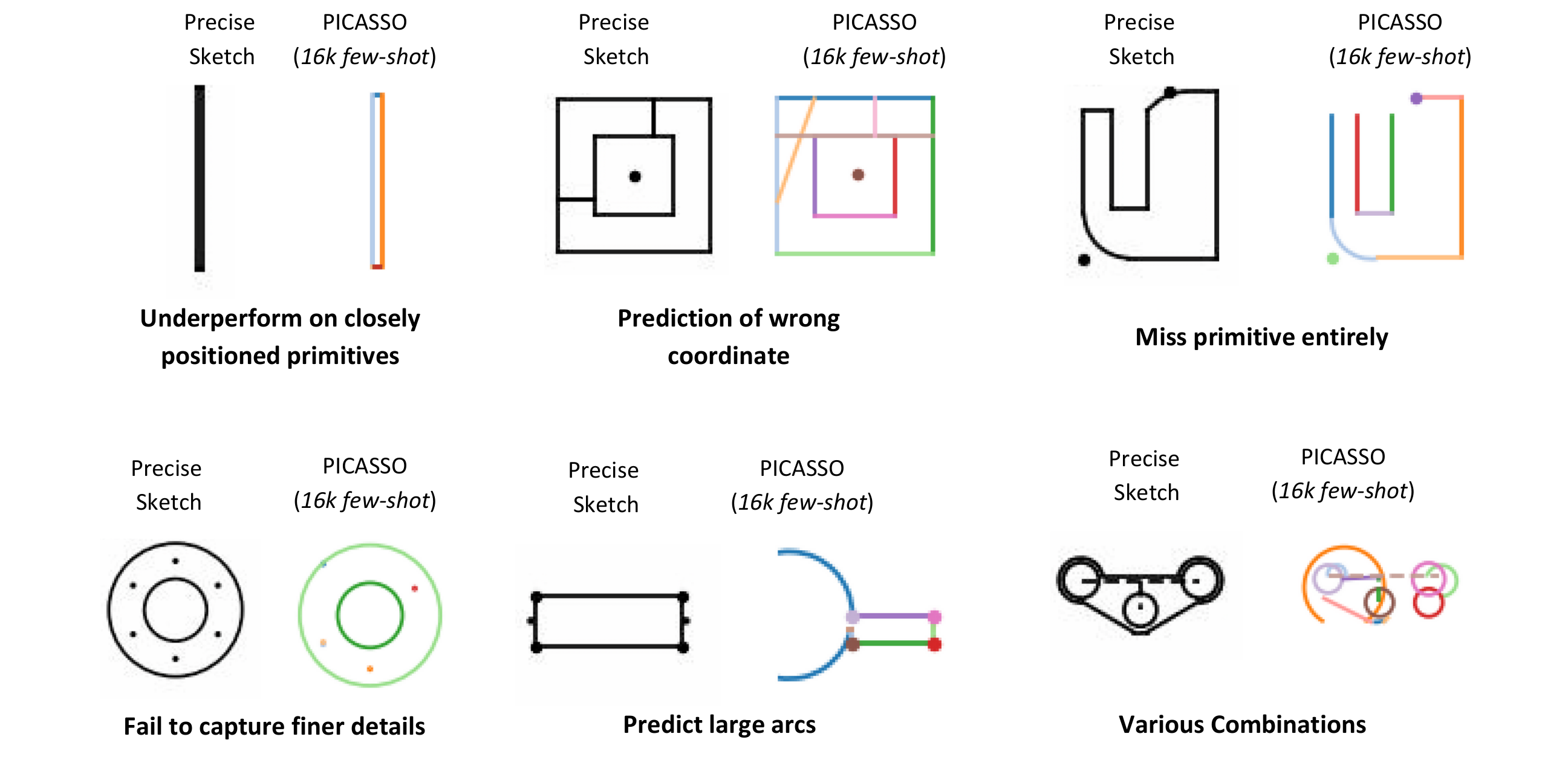}
    \caption{Common failure cases of \texttt{PICASSO} finetuned with parametric supervision on 16k samples.}
    \label{fig:failure_cases}
\end{figure*}

We conduct a qualitative analysis of common failure cases identified for \texttt{PICASSO} finetuned with 16k samples. Figure~\ref{fig:failure_cases} highlights several of these cases, where the model produces suboptimal parameterizations. Notable issues include difficulty handling closely positioned primitives, inaccurate coordinate predictions, missing primitives, failure to capture fine details, overly large arc predictions or combinations of the aforementioned cases.

\clearpage

\section{\texttt{PICASSO} for Feature-based 3D CAD Modeling}

\begin{figure*}[h]
    \centering
	\includegraphics[width=0.8\linewidth]{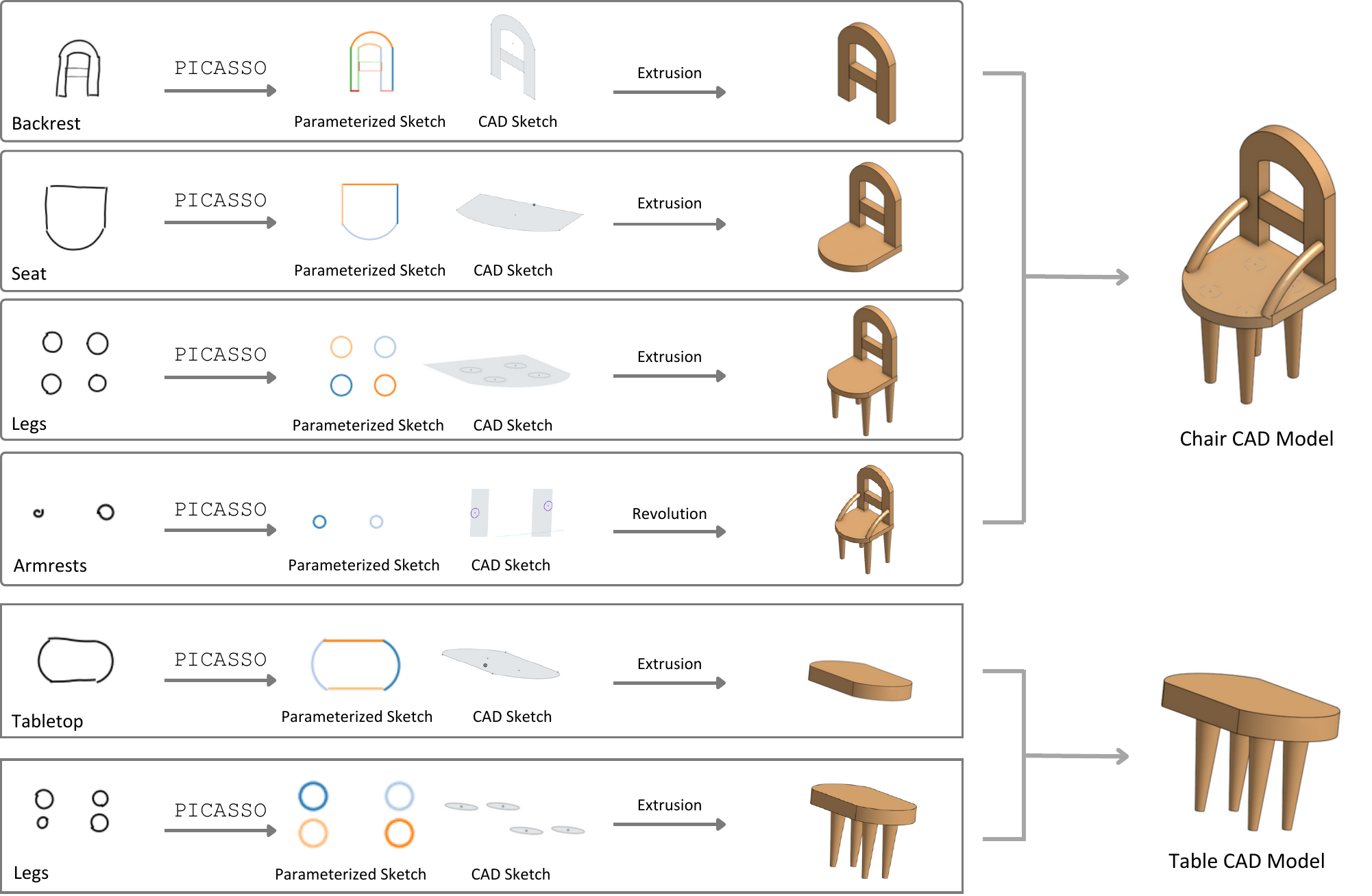}
    \caption{3D CAD modeling from hand-drawn sketches with \texttt{PICASSO}.}
    \label{fig:3d_capabilities}
\end{figure*}

2D CAD sketches are an essential component of feature-based CAD modeling. In Figure~\ref{fig:3d_capabilities}, we demonstrate how \texttt{PICASSO} enables the design of 3D CAD models by parameterizing hand-drawn sketches. These parameterized sketches are then uploaded to Onshape~\cite{Onshape}, where simple extrusions and revolutions are applied. This results in 3D solids that are combined to form the 3D models depicted in the figure.

\end{document}